\documentclass{article}

\usepackage{arxiv}

\usepackage[utf8]{inputenc} 
\usepackage[T1]{fontenc}    
\usepackage{hyperref}       
\usepackage{url}            
\usepackage{booktabs}       
\usepackage{amsfonts}       
\usepackage{nicefrac}       
\usepackage{microtype}      
\usepackage{lipsum}		
\usepackage{graphicx}
\usepackage{natbib}
\usepackage{doi}

\usepackage{times}
\usepackage{subcaption} 
\usepackage{algorithm}
\usepackage{algpseudocode}
\usepackage{multirow}
\usepackage{comment}
\usepackage{amsmath}

\title{A Roadmap Towards Improving Multi-Agent Reinforcement Learning with Causal Discovery and Inference}

\author{ \href{https://orcid.org/0000-0002-9572-3150}{\includegraphics[scale=0.06]{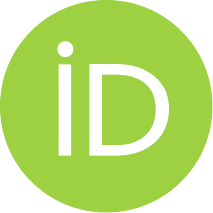}\hspace{1mm}Giovanni Briglia} \\
	Department of Sciences and Method for Engineering\\
	University of Modena and Reggio Emilia\\
	Reggio Emilia, RE 42122 \\
	\texttt{giovanni.briglia@unimore.it} \\
	\And
	\href{https://orcid.org/0000-0001-8921-8150}{\includegraphics[scale=0.06]{orcid.pdf}\hspace{1mm}Stefano Mariani} \\
	Department of Sciences and Method for Engineering\\
	University of Modena and Reggio Emilia\\
	Reggio Emilia, RE 42122 \\
	\texttt{stefano.mariani@unimore.it} \\
    \And
	\href{https://orcid.org/0000-0002-6837-8806}{\includegraphics[scale=0.06]{orcid.pdf}\hspace{1mm}Franco Zambonelli} \\
	Department of Sciences and Method for Engineering\\
	University of Modena and Reggio Emilia\\
	Reggio Emilia, RE 42122 \\
	\texttt{franco.zambonelli@unimore.it} \\
}

\renewcommand{\shorttitle}{Causal MARL: a Roadmap}

\hypersetup{
pdftitle={A Roadmap Towards Improving Multi-Agent Reinforcement Learning with Causal Discovery and Inference},
pdfsubject={cs.AI, cs.MA, cs.LG},
pdfauthor={Giovanni Briglia, Stefano Mariani, Franco Zambonelli},
pdfkeywords={Multi-Agent Reinforcement Learning, Causal Discovery, Causal Inference, Structural Causal Model},
}

\begin{document}
\maketitle

\begin{abstract}
Causal reasoning is increasingly used in Reinforcement Learning (RL) to improve the learning process in several dimensions: efficacy of learned policies, efficiency of convergence, generalisation capabilities, safety and interpretability of behaviour. 
    However, applications of causal reasoning to \emph{Multi}-Agent RL (MARL) are still mostly unexplored. 
    In this paper, we take the first step in investigating the opportunities and challenges of applying causal reasoning in MARL. 
    We measure the impact of a simple form of causal augmentation in state-of-the-art MARL scenarios increasingly requiring cooperation, and with state-of-the-art MARL algorithms exploiting various degrees of collaboration between agents. 
    Then, we discuss the positive as well as negative results achieved, giving us the chance to outline the areas where further research may help to successfully transfer causal RL to the multi-agent setting.
\end{abstract}


\section{Introduction}
\label{sec:intro}

Reinforcement Learning (RL) is now an established approach to let a software agent autonomously learn how to achieve a task in a complex environment requiring sequential decision-making. 
There, the agent learns from the experience of a reward signal obtained via repeated interactions with such environment (\cite{DBLP:books/lib/SuttonB98}). 
Despite its success, there are still open issues that nurture research in RL. 
For instance: 
    how to generalise to novel tasks and/or unseen environments (\cite{DBLP:conf/aaai/MuttiSRCBR23}), 
    how to guarantee that the learned policy adheres to desired safety boundaries (\cite{DBLP:journals/corr/MunosSHB16}), 
    and how to represent the learned policies in a human-interpretable way (\cite{Madumal2020}). 
Different approaches have been proposed in the literature for each of these open issues. 
Here, we focus on those leveraging \emph{causal discovery} and \emph{inference}, that is, the learning and exploitation of cause-effect relationships in the RL system composed by the learning agent and its environment (\cite{10.1093/oxfordhb/9780199399550.013.20}). 

\emph{Causal RL}, in fact, is a thriving field of research aimed at combining \emph{causal models} with RL. 
These models, as formulated in \cite{Pearl1995}, are mathematical frameworks meant to capture cause-effect relationships amongst variables in a system, and to enable qualitative and quantitative reasoning over such relations. 
Their natural link to RL, as well as the benefits they could bring to it, have already been described and demonstrated, for instance by  \cite{DBLP:journals/corr/abs-2307-01452,DBLP:journals/corr/abs-2302-05209}---at least for \emph{single-agent} RL. 
In fact, learning a causal model of a RL system (there including the learning agent and its environment) has been shown to positively impact RL in several ways. 
In particular:  
    consequences of agent actions in given states may be reliably predicted, 
    and actions achieving desired goals reliably planned. 

However, transferring causal RL approaches to \emph{Multi-Agent Reinforcement Learning} (MARL) is still relatively unexplored, as witnessed by \cite{grimbly2021causal}. 
In a multi-agent setting, in fact, RL on its own already becomes much more complicated with respect to the single-agent case. 
The main reason is that multiple agents learning concurrently in a shared environment induce \emph{non-stationarity} of the environment state transitions and an exponentially large \emph{joint action} space~\cite{Du2021}. 
In addition, causal discovery becomes much more difficult, too: the sheer size of the causal model to be learnt, in the number of variables and especially of their links, quickly becomes computationally cumbersome when considering the whole Multi-Agent System (MAS)~\cite{Wang2024}. 
In this paper, we take a step in a broad quantitative investigation of the opportunities and challenges of applying causal reasoning in MARL. 

In particular, we focus on \emph{Structural Causal models} (SCMs) and \emph{interventions} as defined by \cite{DBLP:journals/ijon/Shanmugam01}. 
We develop a conceptual and software framework that augments ``vanilla'' MARL algorithms with causal discovery and incorporates the learnt causal model into the agents' decision-making process. 
This serves to prune the action space of risky actions and nudge the agent towards desired states, via causal inference (i.e.\ prediction and planning). 
We call this framework Causality-Driven Reinforcement Learning (CDRL, for short). 
The goal of the augmentation is to improve the efficacy, efficiency, and safety of the learning process and learnt policy. 

To keep the approach virtually applicable to any MARL environment, we consider for the causal discovery problem the same variables that define any (MA)RL problem, with no additional expert knowledge: the action space, the observation space, and the reward space. 
Accordingly, we only aim to learn the causal relationships linking actions to states, and states to rewards in turn, not all the possible links (e.g.\ between states). 
We call this a \emph{minimal causal model}, or, a causal model of the ``core'' environment dynamics. 
We measure the impact of such a simple form of causal augmentation in three state-of-the-art MARL scenarios increasingly requiring cooperation, and featuring partial observability as well as continuous action and observation spaces. 
These features make the problem of causal discovery particularly difficult, as agents are forced to learn partial causal models of their environment, and interventions are computationally complicated by the continuous nature of action and state variables. 
As the vanilla MARL algorithm to be causally augmented, we choose three state-of-the-art MARL algorithms exploiting various degrees of collaboration between agents (e.g.\ parameters sharing, value function decomposition, joint action space factorisation, etc.). 
Finally, we discuss the positive as well as negative results achieved, giving us the chance to outline the areas where further research may help to successfully transfer causal RL to the multi-agent setting. 

The paper is structured as follows: 
    Section~\ref{sec:back} provides background information and related works; 
    Section~\ref{sec:causality_driven_rl} describes our approach to causal augmentation of MARL algorithms; 
    Section~\ref{sec:eval} evaluates our approach in state-of-the-art scenarios to provide quantitative ground for the research roadmap discussed in Section~\ref{sec:discussion}; 
    finally, Section~\ref{sec:conc} concludes the paper. 

\section{Background and Related Works}
\label{sec:back}

In this section, we quickly recall the very basics of MARL and causal reasoning, 
by focusing on SCMs 
and on the additional challenges that concurrent learning brings along in MARL with respect to single-agent RL, 
then we overview the state of the art for both causal RL and causal MARL. 

\subsection{From Causal RL to Causal MARL}

\emph{Causal models} are a particularly desirable kind of model of (causal) dependencies between variables in a domain (e.g.,\ an RL environment), as they enable to disambiguate true causation between state variables from mere correlation (\cite{DBLP:books/crc/chb/Pearl14}). 
They are known to capture relationships and express reasoning that is out of reach for purely statistical machine learning approaches (\cite{DBLP:journals/cacm/Pearl19}), and their connections with and usefulness to RL research has already been motivated, for instance by \cite{DBLP:journals/pieee/ScholkopfLBKKGB21,10.1093/oxfordhb/9780199399550.013.20}. 
%
    In the context of RL, a causal model helps understand (and thus predict) how actions (cause) may change the state of the environment (effect), and, in turn, how to plan actions (cause) to achieve a desired state (effect). 

A Structural Causal Model (SCM) is a specific formalism to represent causal models that expresses the causal relationships among variables using a Directed Acyclic Graph (DAG) alongside a set of structural equations. 
In an SCM, each node in the DAG corresponds to a variable in the system, and directed edges between nodes indicate direct causal influences. 
The structural equations specify how each variable is causally determined by its parent variables in the graph---quantitatively. 

\emph{Causal discovery} is the process of learning such models based on data, possibly interventions, and identifiability conditions (see \cite{DBLP:journals/corr/abs-1202-3757}). 
In the context of RL, the data are, for instance, the action-state-reward trajectories produced by the agent while acting in the environment. 
\emph{Interventions} are a fundamental mechanism for causal discovery, and also for causal inference, that is the analysis of how variables' values change given others. 
They involve deliberately manipulating certain variables within a system (by changing their values) to observe their effects on others, thereby unravelling the causal relationships at play~\cite{doi:10.1080/01621459.2012.754359}. 
    In the case of the model of an RL environment, this translates to being able to plan what actions are needed to bring about a desired effect (i.e.\ a state of the environment) and to predict the effect of actions carried out in a given state. 
Interventions have been formalised with the $do$ operator~\cite{DBLP:conf/uai/HuangV06}, which enables to express statements like $P(Y \mid do(X = x))$ to represent the probability distribution of variable $Y$ given an intervention setting variable $X$ to value $x$. 

These models (as well as other frameworks for causal reasoning) can be used in both single-agent and multi-agent RL 
with the purpose of improving the learning process and policies in a number of directions 
    (e.g.\ efficacy, efficiency, safety, generalisation, interpretability, etc.). 
However, multi-agent settings are notably more complex than single-agent ones, 
both for the typical additional challenges of MARL with respect to single-agent RL, 
and for the inherent complexity of causal discovery (first and foremost), 
and causal inference when many variables and potential causal relationships are at play. 
For the former, the issues are mostly about the induced non-stationarity of environment state transitions due to concurrent learning of agents, credit assignment, and equilibrium selection issues. 
For the latter, most causal discovery algorithms entails some sort of conditional independence testing and/or experimental design routine that scales poorly in the number of nodes and/or causal relationships. 
Also, when multiple agents want to learn a causal model of their relationships with the environment, they would likely need to also model other agents' influences. 

The next Subsection overviews the state-of-the-art in this regard, as a confirmation that causal MARL is still relatively unexplored yet. 

\subsection{Causal (MA)RL}

Before overviewing the landscape of causal RL for both single-agent and multi-agent systems, a premise is necessary. 
Lots of literature uses causal terminology and claims to perform causal reasoning to some extent, and there are many actual methods to support causal reasoning nowadays. 
However, in this paper, we are mostly concerned with \emph{explicit} causal reasoning frameworks, especially those grounded in \emph{Structural Causal Models} (SCMs) and the notion of intervention to separate association from causation in the causal ladder~\cite{DBLP:journals/ijon/Shanmugam01}---as popularised first and foremost by Judea Pearl. 
The reason for adopting this specific view, albeit perhaps narrow, is that it is mathematically solid, has been made operational by many works (starting from the do-calculus~\cite{DBLP:conf/uai/HuangV06}), and fosters an explicit representation of causal models---that is, amenable of rigorous and mechanic manipulation. 
Other approaches, instead, cannot claim all of these nice properties. 
For instance, many approaches adopt \emph{deep learning} to learn causal relationships between variables in a \emph{latent space}, such as \cite{DBLP:conf/ijcai/Hosoya19,yao2018direct,le2024multi,pina2023discovering}.  
Other use ad-hoc or statistical notions underlying causal relationships and especially to quantify causal effects, for instance based on information-theoretic and/or statistical measures such as conditional mutual information. 
Examples are \cite{DBLP:journals/inffus/YangYZK23,DBLP:conf/icml/LiK0LCF0022,DBLP:journals/nn/LiWSWJWDYTH24,reizinger2024identifiable} 
Some approaches use both, such as \cite{DBLP:journals/ieeejas/JiangLWDS24,liesen2024discovering}. 

Narrowing down our attention to the learning (discovery) and/or exploitation (inference) of explicit models of cause-effect relationships adhering to the premise disclosed above, single-agent causal RL is more mature, whereas causal MARL is less explored---especially for causal discovery. 
%
%
The majority of contributions focus on \emph{causal inference} or induction (as defined by \cite{genewein2020algorithms}), that is, exploiting known causal structure to estimate effects of variables on others. 
An example is the work by \cite{DBLP:conf/nips/SeitzerSM21}, where conditional mutual information is used as a basis to define a measure of ``causal influence'' enabling a learning agent to detect when it is in control of some environment dynamics. 
With such information, the agent can improve exploration. 
\cite{DBLP:conf/aaai/MuttiSRCBR23,mutti2023exploiting}, instead, aim to exploit causality to improve generalisation abilities of an RL agent. 
They define causality as the common structure across MDPs and invariant in time, such as the law of motion according to physics. 
Hence they target a set of environment assumed to share such common structure, and show how learning the causal structure by sampling different environments makes the agent \emph{generalise} to unseen environments---but with the same common structure. 
%
For \emph{causal discovery}, instead, \cite{lu2021nonlinear} propose invariant Causal Representation Learning (iCaRL), an approach that enables out-of-distribution (OOD) generalization not only for RL but potentially for machine learning in general. 
\cite{DBLP:conf/nips/KocaogluSB17} adopts an experimental design approach to propose two causal discovery algorithms. 

Application of causal reasoning in \emph{multi-agent} RL (MARL) has been envisioned by \cite{grimbly2021causal} but is significantly less explored to date compared to single-agent RL. 
Here, causal inference is more common than discovery, and mostly used to adjust individual rewards based on detected causal relationships, with the goal of enhancing credit assignment and promote cooperation. 
For instance, \cite{DBLP:journals/corr/abs-2312-03644} leverage causal modelling in offline multi-agent reinforcement learning by utilizing a Dynamic Bayesian Network to capture causal relationships between agents' actions, states, and rewards. 
    This approach decomposes team rewards into individual contributions, facilitating precise and interpretable credit assignment that promotes cooperation and improves policy learning. 
Another recent contribution, by \cite{DBLP:conf/aaai/DuYZYCW24}, introduces a MARL approach for identifying situation-dependent causal influences between agents. 
    By detecting states where one agent's actions significantly impact another, they promote more effective coordination. 
    Their MACGM framework formalizes the CGM paradigm in a multi-agent setting and incorporates an intrinsic reward to enhance awareness of causal influences among agents. 

\section{Proposed Causal Augmentation}
\label{sec:causality_driven_rl}

\begin{figure}[!t]
    \centering
    \includegraphics[width=.6\columnwidth]{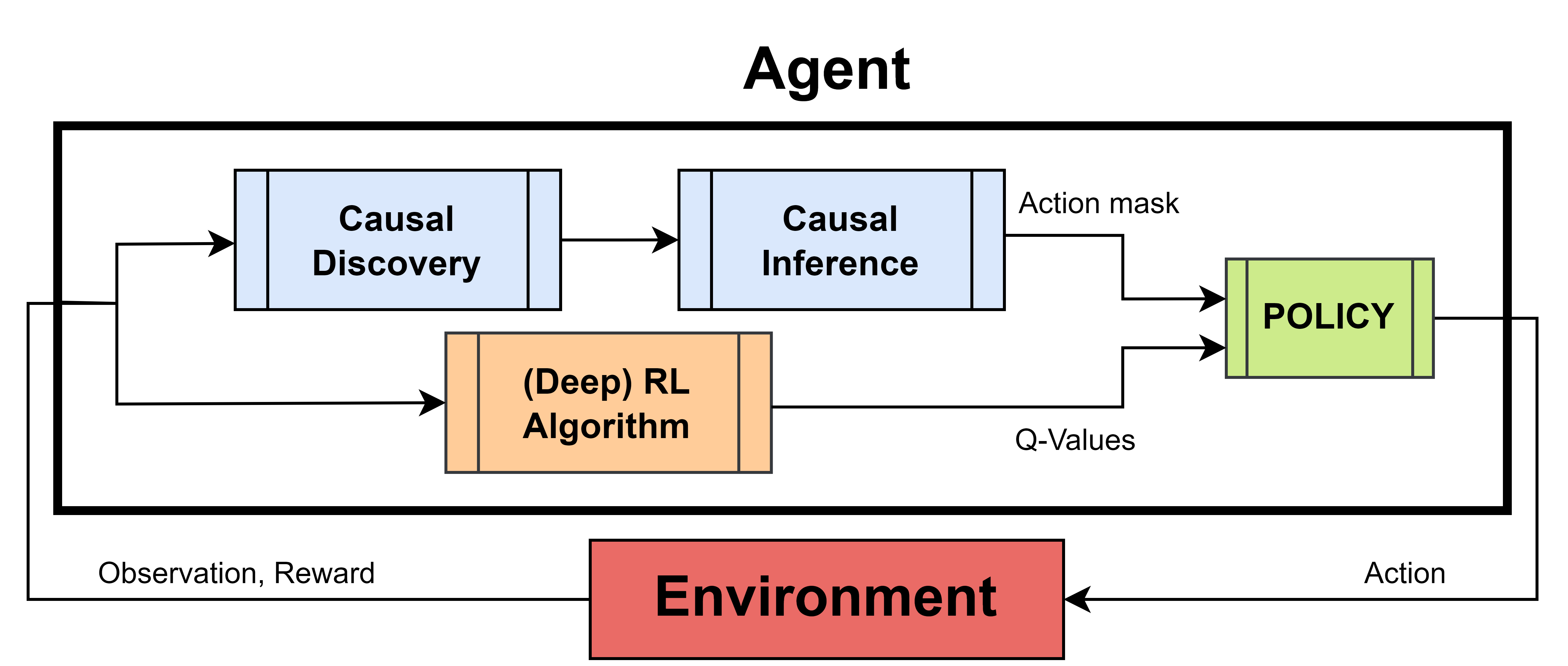}
    \caption{Causal augmentation architecture: the agent interacts with the environment first to learn a minimal causal model, then to learn an action policy. There, in the action selection step, causal inference modulates the action space by acting as a filter (action mask).}
    \label{fig:causal_rl_frameworks}
\end{figure}

Our proposed causal augmentation amounts to undertake two key consecutive tasks---depicted in Figure~\ref{fig:causal_rl_frameworks}: 
    (1) \emph{causal discovery} of a (minimal) causal model that relates actions, states, and the reward signal, 
    and (2) leveraging this knowledge to provide the agent with an action filter (or, mask) via \emph{causal inference}. 
%
As the general MARL problem formulation to apply our proposed causal augmentation, we take as a reference \emph{Partially Observable Markov Games} (POSGs), following \cite{bettini2023heterogeneous, DBLP:conf/nips/LiuS022}. 
All the experimental scenarios used for our investigation in Section~\ref{sec:eval} abide to this standard formulation. 
It is worth noting that the simplicity of our proposed causal augmentation enjoys a few desirable properties: 
it is algorithm-agnostic, meaning that it can be applied to augment virtually any MARL (and RL) algorithm; 
can be used with tabular methods (such as Q-learning and other TD-learning techniques) as well as non-tabular ones using any kind of (deep) neural networks as function approximations; 
it is compatible with both continuous and discrete action spaces. 

\subsection{Causal Discovery in MARL}


We treat the agent's actions on the environment as interventions on the system's variables. 
Thus, the data in input to the causal discovery process consists of the agent's trajectory, represented by a sequence of observation-action-reward triplets. 
The output is the learned structural causal model, composed by a DAG, and the Conditional Probability Distributions (CPDs) of each node therein, used as the structural equations of the model. 
This enables inferences such as action planning and future state prediction (as described in the next subsection). 
%
%
Causal discovery proceeds as follows: 
\begin{enumerate}
    \item each learning agent repeatedly acts in the environment by following any arbitrary policy (e.g.\ simply $\epsilon$-greedy Q-learning); 
    \item the agent trajectories are collected; 
    \item the PC algorithm by \cite{DBLP:books/daglib/0023012} as implemented in \cite{DBLP:journals/jmlr/ZhengH0RGCSS024} is fed with such trajectories, \emph{independently for each agent}, and learning of the causal model starts; 
    \item in the process, the DAG is constrained to have the reward variable as a children of any (learnt) subset of the action and/or state variables, and not to be a parent of any other node. 
\end{enumerate}
As the specific data structure representing such a model, we rely on a (Causal) Bayesian Network (CBN), whose DAG corresponds to the causal DAG, and whose CPDs play the role of the causal structural equations. 


\subsection{Causal inference in MARL}

At this stage, the learnt causal model is exploited to perform \emph{causal inference}, that is, predicting the effects of actions on observable states, and planning actions to achieve desired rewards. 
The input to this stage is the current observation, as in any RL algorithm, and the output is a pool of admissible actions, filtered according to causal inference as follows. 

\begin{enumerate}
    \item At each action selection step, we manipulate the CBN by conditioning on the observation variables in the Markov Blanket~\cite{DBLP:journals/jmlr/PelletE08} of the reward variable, that is, we set their values to the one in the current observation. 
    \item Then, we perform interventions on each action variable, by iteratively assigning to each of them every admissible value---exploiting Pearl's $do$ operator. 
    This process enables to compute the probability distribution of the reward variable (\emph{prediction}), conditioned on the observations, and hypothesizing a given action (\emph{planning})---the intervention. 
    \item Based on these distributions, we determine a \emph{likability score} for each action, by weighting the rewards with their respective probabilities. 
    \item Upon these scores we compute an action mask based on a pluggable filtering strategy. 
    For instance, for the assessment in Section~\ref{sec:eval}, we look at percentiles: we filter out actions with scores below the 25th percentile of the whole batch of scores, and, in case there are actions above the 75th percentile, we filter out also any other action (to nudge the agent to choose amongst these best ones). 
    \item The mask is applied to the action pool, and the MARL algorithm can select amongst them. 
\end{enumerate}
The entire process is outlined in Algorithm \ref{alg:causal_forward_pass}.

\begin{algorithm}
    \caption{Causal inference stage of the causal augmentation, executed at each agent-environment interaction (or in batches, by replacing $obs_{cur}$ with the batch of observations)}\label{alg:causal_forward_pass}
    \begin{algorithmic}
        \Require $CBN$: Causal Bayesian Network, $obs_\text{cur}$: current observation
        \begin{enumerate}
            \item Condition nodes in the Markov Blanket of the reward node with $obs_{cur}$; 
            \item Compute the reward inference data structure outlined in Equation~\ref{eq:inference}; 
            \item Compute the likability scores for each action; 
            \item Compute the action mask based on an arbitrary strategy (e.g. percentiles); 
            \item Apply the obtained action mask to the action select step of the (MA)RL algorithm. 
        \end{enumerate}
    \end{algorithmic}
\end{algorithm}

To further clarify the approach, the following data structure is a convenient representation of the output of step §2: 
\begin{equation}\label{eq:inference}
\text{reward inference} = 
\begin{cases}
    do(a_1) | obs_{cur}: \{ r_1^1: \mathcal{P}(r_1^1), \dots, r_1^k: \mathcal{P}(r_1^k) \}, \\
    \vdots \\
    do(a_n) | obs_{cur}: \{ r_n^1: \mathcal{P}(r_n^1), \dots, r_n^m: \mathcal{P}(r_n^m) \}
\end{cases}
\end{equation}
where $a_{1...n}$ denotes an action value, $obs_{cur}$ denotes the current observation, $r_{1...n}^{1...}$ denotes the set of reward values seen for action $n$, $|$ denotes conditioning, and $\mathcal{P}$ denotes a probability distribution.
 
Notice that such a formulation hints at discrete action and reward spaces just for the sake of clarity. 
In the case of continuous spaces, any technique for carrying out causal interventions on continuous variables can be applied---e.g., discretisation, soft interventions~\cite{DBLP:conf/nips/KocaogluJSB19} working with distribution shifts (see the roadmap in Section~\ref{sec:discussion}). 


\section{Application to MARL}
\label{sec:eval}

Within the framework described above, we want to assess whether, and to what extent, our proposed causal augmentation is able to improve efficacy, efficiency, and/or safety of the MARL learning process and resulting policy. 
To this end, a set of ``vanilla'' MARL algorithms are augmented with causal discovery, and then incorporate the learnt causal model into the agents' decision-making process. 
In fact, the model is used to perform causal inference (prediction and planning) on variables representing states, actions, and the reward signal, to obtain an \emph{action filter}. 
Such filter serves to modulate the action space available to the agent dynamically, by pruning risky actions, while nudging the agent towards those more likely to lead to desired states (e.g.\ with higher expected rewards).  
By comparing the performance of the causally augmented version of the algorithms with their vanilla versions, in accomplishing tasks requiring different levels of cooperation, we can discuss the opportunities and challenges of applying causal reasoning to MARL problems. 

Accordingly, this section aims to address the following research questions (RQs): 
\begin{enumerate} 
    \item Can the augmentation of a MARL algorithm with a minimal causal model of the environment dynamics improve the learning process in terms of efficacy, efficiency, and safety? \textbf{(RQ1)}
    \item If improvements are observed, which combinations of tasks and algorithms exhibit them the most? \textbf{(RQ2)} And why? 
    \item If not,  which combinations of tasks and algorithms exhibit the worst difference in performance? \textbf{(RQ3)} And why?
\end{enumerate}
In the following, we provide quantitative measures to objectively respond to the first part of each question. 
Then, next section speculates on the ``why'' part, grounded on these quantitative results. 

\subsection{Experimental Settings}

\begin{figure}[!t]
    \centering
        \begin{subfigure}[]{.2\columnwidth}
            \includegraphics[width=\columnwidth]{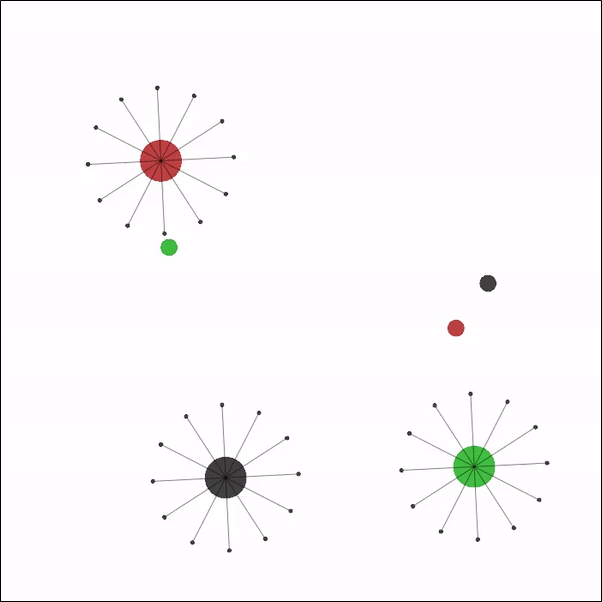}
            \caption{Navigation}
            \label{fig:navigation_env}
        \end{subfigure} 
        \begin{subfigure}[]{.2\columnwidth}
            \includegraphics[width=\columnwidth]{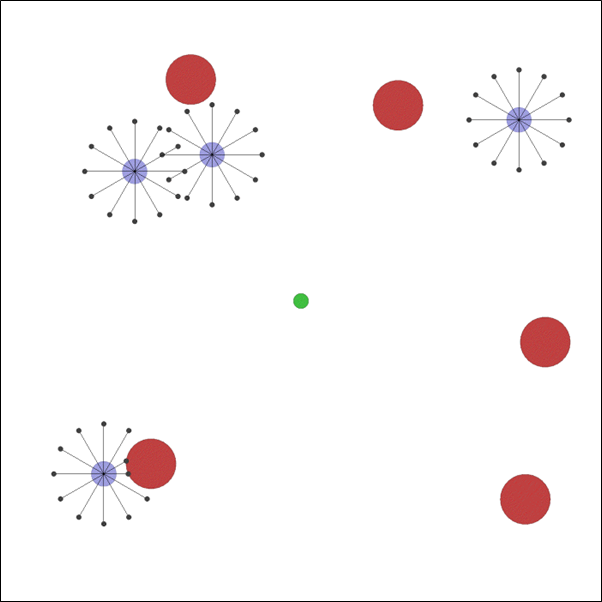}
            \caption{Flocking}
            \label{fig:flocking_env}
        \end{subfigure} 
        \begin{subfigure}[]{.2\columnwidth}
            \includegraphics[width=\columnwidth]{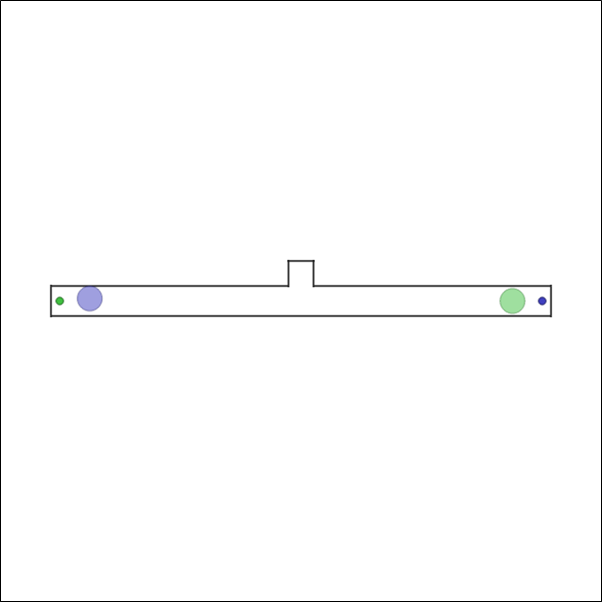}
            \caption{Give-Way}
            \label{fig:give_way_env}
        \end{subfigure}
    \caption{Experimented multi-agent tasks from the VMAS simulator by \cite{DBLP:conf/dars/BettiniKBP22}.}
    \label{fig:VMAS_envs}
\end{figure}

\textbf{Environments.} We use the VMAS simulator by \cite{DBLP:conf/dars/BettiniKBP22} that offers a variety of multi-agent environments. 
In particular, we consider three task environments chosen as they require different degrees of \emph{cooperation} to the agents: navigation, flocking, and give-way---increasingly requiring cooperation. 
All of them feature partial observability, continuous observation and action spaces, and dense rewards. 
Figure \ref{fig:VMAS_envs} provides a visual representation of these environments. 
The Navigation scenario (Figure \ref{fig:navigation_env}) requires each agent to reach its own colour-coded goal location while avoiding collisions with others. 
In Flocking (Figure \ref{fig:flocking_env}) agents must coordinate to circle around a given goal location as a cohesive group while evading obstacles. 
Finally, in Give-Way (Figure \ref{fig:give_way_env}) agents need to walk down a narrow hallway by letting each other through. 

\noindent \textbf{RL Algorithms.} We take off-the-shelf algorithms from the BenchMARL framework by \cite{DBLP:journals/jmlr/BettiniPM24} and augment them with our causal discovery and inference pipeline. 
Algorithms are chosen based on their support to and exploitation of agent collaboration, similar to environments: 
    Independent Q-Learning (IQL) is non-cooperative (agents learn independently, see \cite{Tan1997MultiAgentRL}), 
    Value Decomposition Networks (VDN) factorises the joint action space based on a shared reward function, hence promotes cooperation implicitly (\cite{sunehag2017value}), 
    and Qmix explicitly shares learning parameters (\cite{DBLP:journals/jmlr/RashidSWFFW20})---hence, is the most collaborative. 
The corresponding causal augmentations are called in the following CausalIQL, CausalVDN, and CausalQmix---their architecture is detailed in Appendix~\ref{app:architecture_causal_benchmarl}. 
%


\noindent \textbf{Metrics.} To quantitatively support the roadmap outlined in Section~\ref{sec:discussion}, we tracked several performance metrics during our experiments. 
They belong to two categories: 
    \emph{scenario-independent} metrics, reported for each scenario, and \emph{scenario-dependent} ones, reported only for specific ones. 
    The former are all normalized in [0,1] and include the mean, median, and Inter-Quartile Mean (IQM, recommended by \cite{DBLP:conf/nips/AgarwalSCCB21}) of the reward achieved after training, during the evaluation stage. 
    They also include the optimality gap, that represents how far the results are from the optimum (i.e.\ 1)---displayed with both its mean and standard deviation. 
    Scenario-dependent metrics, instead, include, for instance, how many collisions happened during training, and the distance between agents and their goal. 
All the metrics are averaged over 10 independent runs, following the recommendations by \cite{DBLP:conf/nips/GorsaneMKDSP22}.

\noindent \textbf{Technical Considerations.} 
Performing interventions on continuous variables in a computationally tractable way is still an open issue in causal discovery and inference literature (see \cite{Wang2024}). 
The formal correctness and practical performance of the few approaches proposed, such as in \cite{DBLP:conf/nips/LorchRSK21,schindl2024incremental,schweisthal2024learning,Wiedermann2022}, is still to be fully assessed and widely agreed upon. 
In the roadmap discussed in Section~\ref{sec:discussion} we provide a brief overview of the most promising research directions available to date. 
Given these considerations, in this preliminary work we decided to discretize observations before feeding it to the causal machinery. 
We thus conducted a sensitivity analysis to ensure that such a discretisation does not catastrophically impact performance---see Appendix~\ref{app:sensitive}. 
\subsection{Results}

In this section we report the results obtained in our experiments, as measured by our described metrics. 
We recall that the goal of this assessment is not to prove that our causal augmentation approach is sufficient or best in class to improve learning in any MARL scenario, but to provide food for thought regarding opportunities and challenges raised by the approach---discussed in Section~\ref{sec:discussion}. 

\begin{figure}[!t]
    \centering
    {%
        \begin{subfigure}[]{.32\columnwidth}
            \includegraphics[width=\columnwidth]{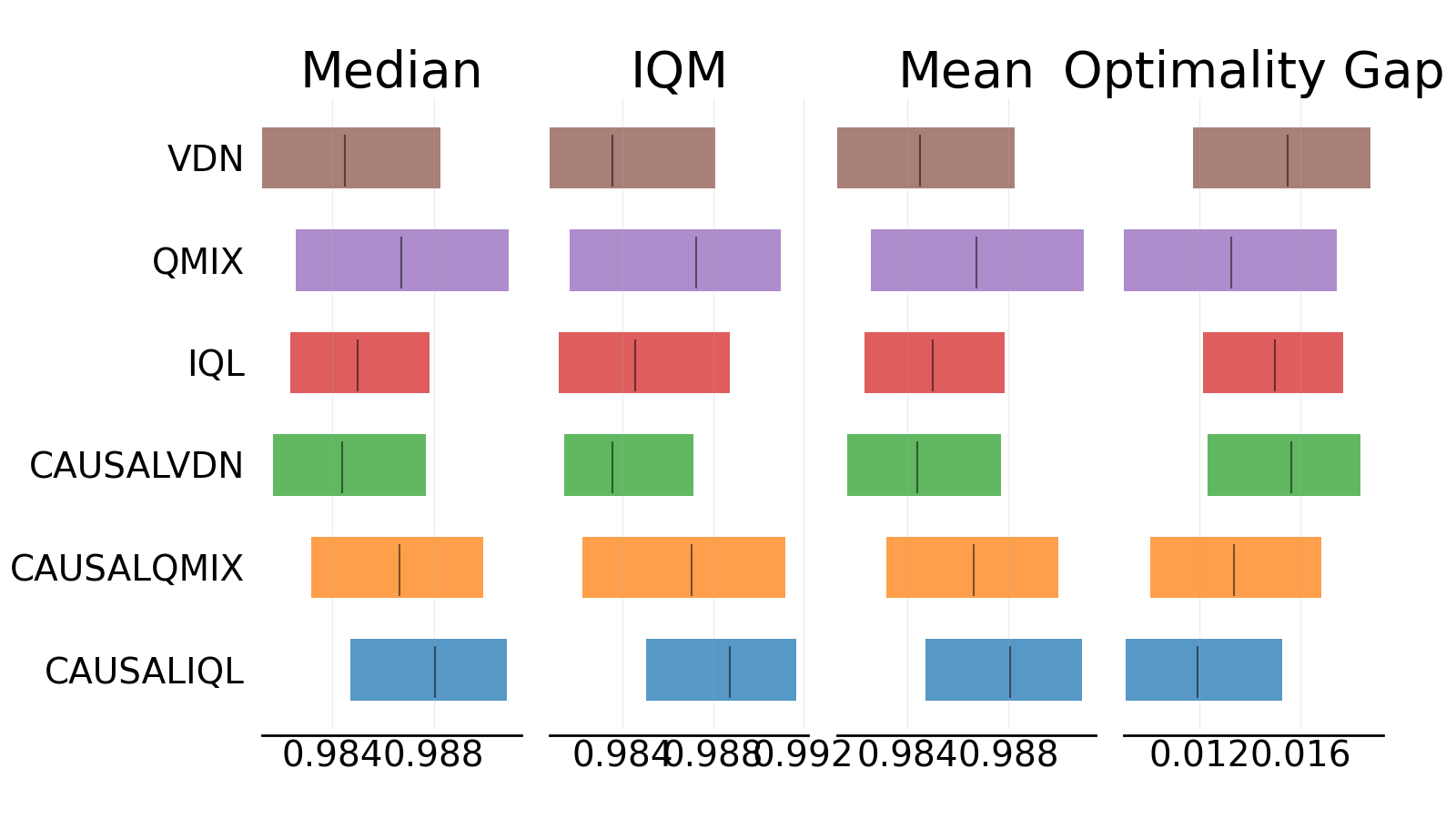}
            \caption{Navigation}
            \label{fig:navigation_aggregate_scores}
        \end{subfigure}
        \begin{subfigure}[]{.32\columnwidth}
            \includegraphics[width=\columnwidth]{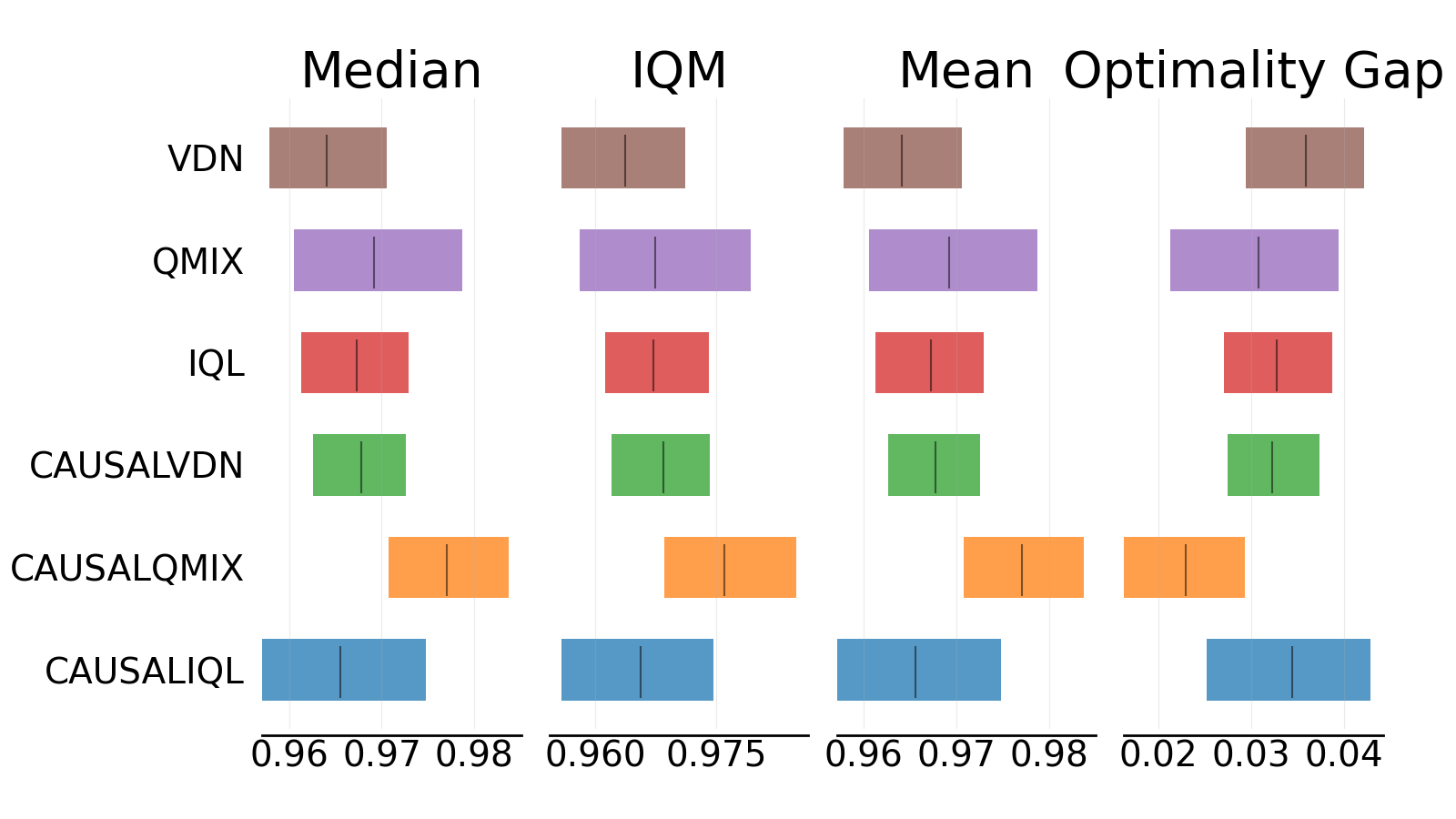}
            \caption{Flocking}
            \label{fig:flocking_aggregate_scores}
        \end{subfigure}
        \begin{subfigure}[]{.32\columnwidth}
            \includegraphics[width=\columnwidth]{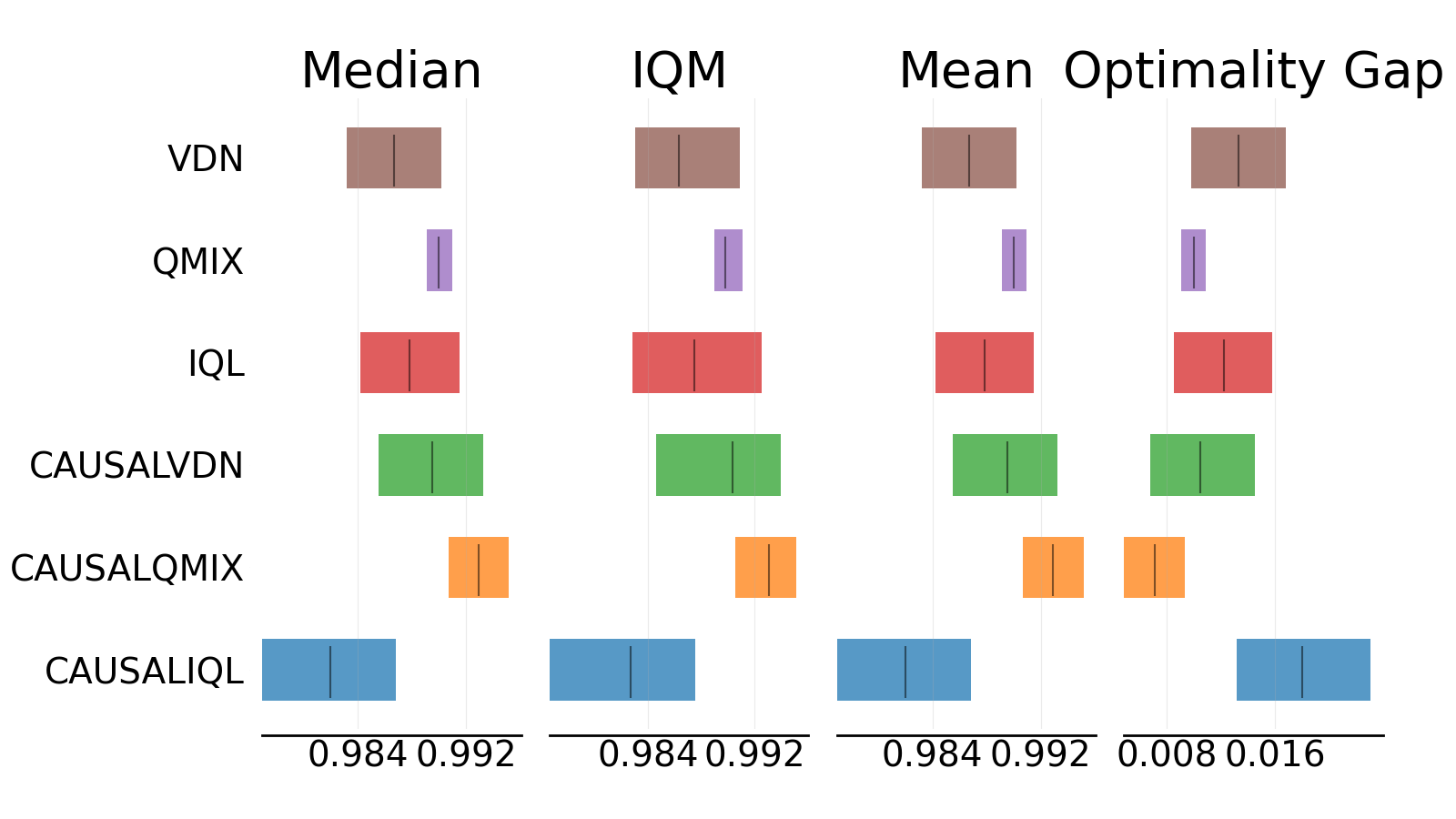}
            \caption{Give-Way}
            \label{fig:giveway_aggregate_scores}
        \end{subfigure}
    }
    \caption{Aggregate scores of median, IQM, mean of normalized reward (all the higher -- to the right -- the better), and optimality gap (the lower -- to the left -- the better), for each algorithm, and for each task.}
    \label{fig:aggregate_scores}
\end{figure}

Figure~\ref{fig:aggregate_scores} shows the scenario-independent metrics for each MARL algorithm (causally augment and vanilla) and for all the scenarios. 
The vanilla versions already perform quite well, nearly reaching optimality in all the scenarios. 
This suggests that further improving such performance with causal augmentation is challenging per se. 
However, there are improvements in specific cases (\textbf{RQ1},\textbf{RQ2}):
    CausalIQL improves IQL in the navigation task, 
    CausalVDN and CausalQmix improve the corresponding vanilla versions in both the flocking scenario, and the give-way scenario. 
Notice that, except for CausalVDN in give-way, also the standard deviation of the metrics is reduced. 
In the other cases, causal augmentation does not affect performance notably or even worsen them, as in the case of CausalIQL in the flocking and give-way tasks (\textbf{RQ3}). 
Recall that tasks and algorithms have different degress of ``cooperativeness'' required/leveraged. 
This could explain variance of these mixed results. 
In particular, IQL is an independent learning algorithm, hence agents do not cooperate in any way during training (on the contrary, the hinder each other with induced non stationarity). 
Thus IQL naturally struggles in highly cooperative tasks, such as flocking and give-way. 
Here, the causal augmentation cannot provide much help, especially cause the naive approach exploited here does not leverage agent cooperation (not in discovery, nor in inference). 

To investigate further and provide all the possible evidence to the roadmap coming in Section~\ref{sec:discussion}, we now turn to some scenario-dependent metrics. 

\begin{figure}[!b]
    \centering
    {%
        \includegraphics[width=0.5\columnwidth]{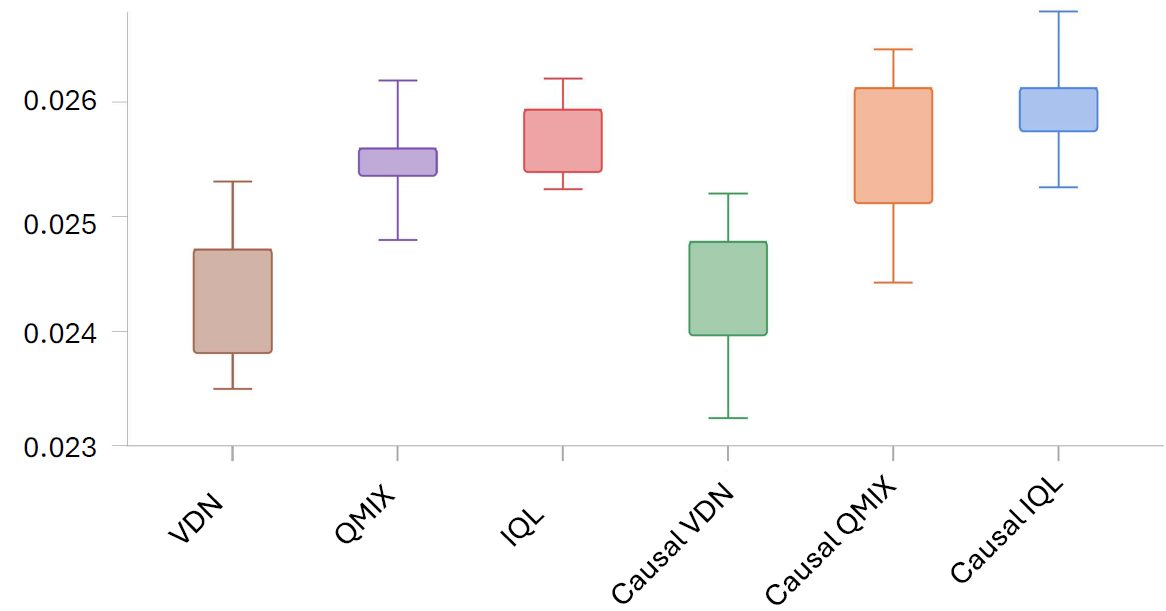} 
    }
    \caption{Data distribution for metric $pos_{rew}$ measuring the closeness of agents to goal in the Navigation task (the higher the better).}
    \label{fig:navigation_plots} 
\end{figure}

In Figure~\ref{fig:navigation_plots} we zoom-in on a reward metric specific to the navigation scenario, $pos\_rew$, that measures how close each agent is to its own goal (the higher the better). 
Here, causal augmentation consistently improves the vanilla baselines both in average performance and variance (\textbf{RQ2}), except for Qmix (\textbf{RQ3}). 
Again, the reason can be found under the cooperation perspective: while navigation is the least cooperative scenario tested, Qmix is the most collaborative algorithm. 
Hence, Qmix alone is able to successfully learn cooperative policies, and the causal augmentation (independently applied to each agent) cannot help further. 

\begin{figure}[!t]
    \centering
    {%
        \begin{subfigure}[]{.45\columnwidth}
            \includegraphics[width=\columnwidth]{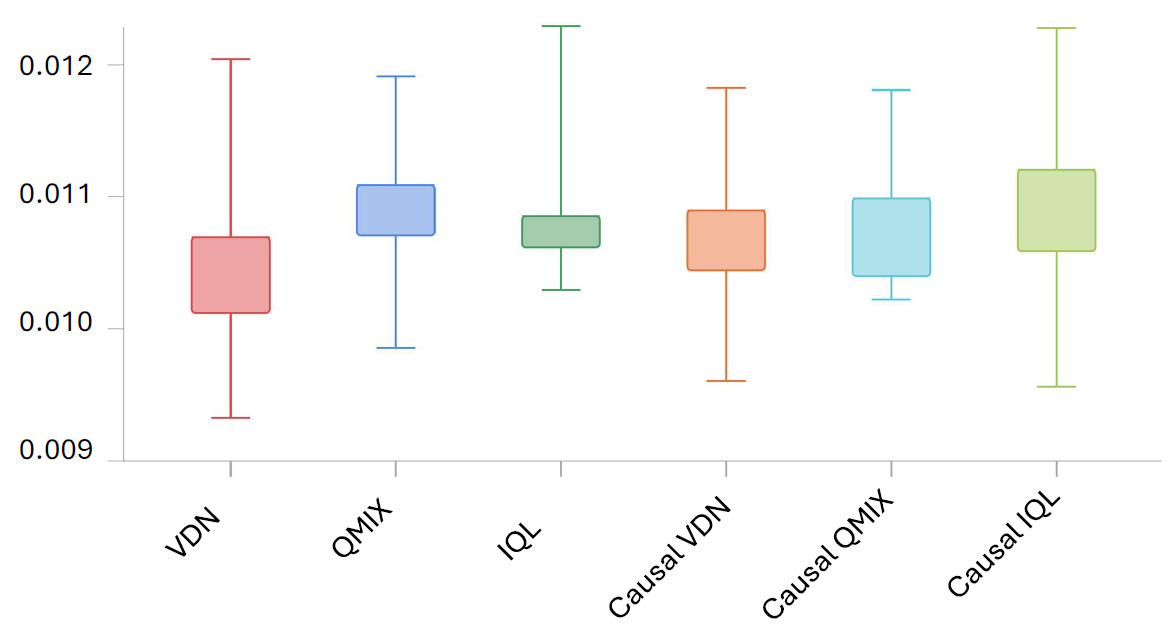}
            \caption{Distance reward (higher is better)}
            \label{fig:flocking_agent_distance_rew_bars}
        \end{subfigure}\hspace{0.02\textwidth} 
        \begin{subfigure}[]{.45\columnwidth}
            \includegraphics[width=\columnwidth]{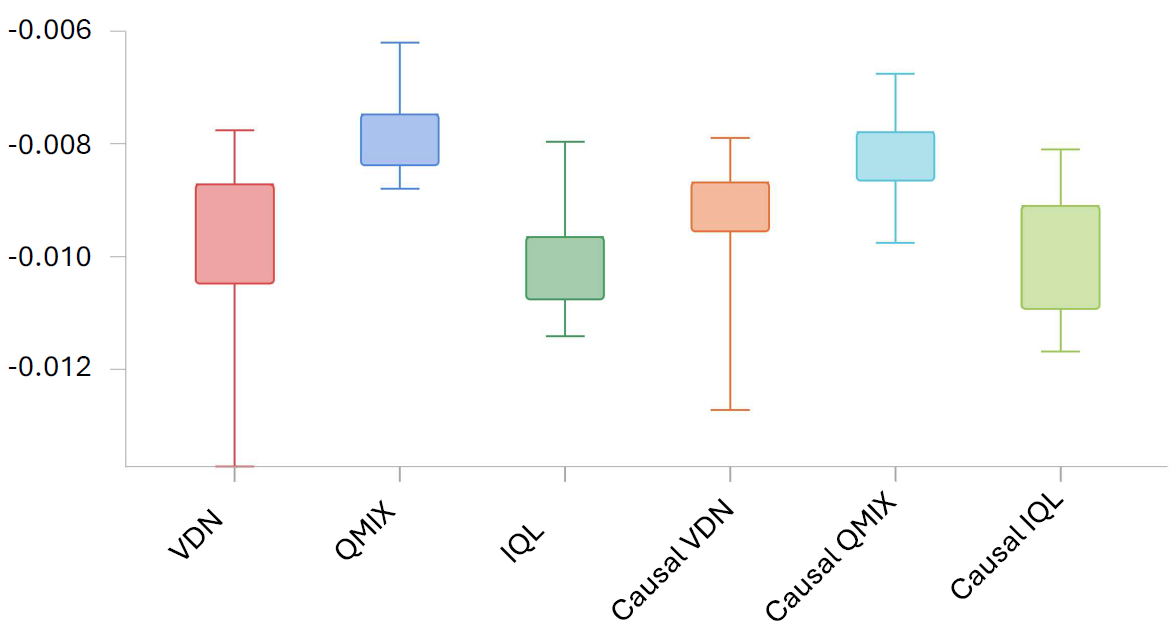}
            \caption{Collisions (higher values are better)}
            \label{fig:flocking_agent_collision_rew_bars}
        \end{subfigure}
    }
    \caption{Data distribution for reward metrics considering the inverse of the distance to goal and the collisions between agents, in the Flocking task.}
    \label{fig:flocking_plots} 
\end{figure}

Figure~\ref{fig:flocking_plots} instead zooms-in on scenario-dependent metrics for the flocking task: 
    the $agent\_distance\_rew$ metric, similar to the $pos\_rew$ in the navigation scenario, 
    and the $agent\_collision\_rew$ metric, which tracks the negative rewards agents accumulated due to collisions.  
We observe mixed results (\textbf{RQ2},\textbf{RQ3}): 
    CausalVDN slightly improves both metrics, although collisions only in variance; 
    CausalQmix only improves collisions, 
    and CausalIQL does not really improve either one, as the slight improvement in distance reward data distribution is balanced by increased variance. 
The same aforementioned considerations about cooperativeness do apply here. 
A particularly relevant result, though, is that collisions are better handled by the causal augmentation. 
This is the result of focussing on ``core'' environment dynamics, and on trying to avoid ``catastrophic'' actions. 
Hence, causal augmentation can open up novel ways of pursuing safety in RL (training and learnt policies). 

\begin{figure}[!b]
    \centering
    {%
        \includegraphics[width=0.5\columnwidth]{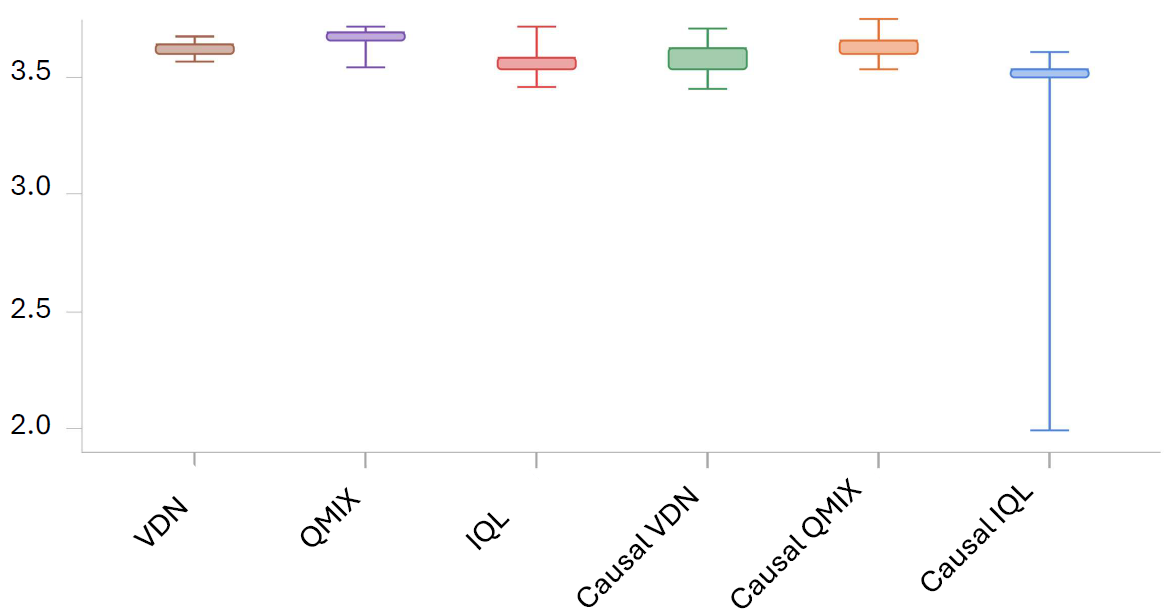} 
    }
    \caption{Give-Way task, $episode\_reward\_mean$ metric (higher is better).}
    \label{fig:giveway_plots} 
\end{figure}

Finally, in Figure~\ref{fig:giveway_plots} we zoom-in on the mean episodic reward of the give-way task, that confirms the exceptional difficulty of CausalIQL in keeping up with all the other approaches, even its vanilla version (\textbf{RQ3}). 
Not surprisingly, the causal augmentation struggles in the ``worst combination'' of task/algorithm: independent learning (IQL) exploited in the most highly collaborative scenario (Give-Way). 

\section{Roadmap towards Causal MARL}
\label{sec:discussion}

Given the preliminary results presented in previous section, 
we can now speculate on the promising research directions opened, and open issues raised, 
by attempting to apply causal discovery and inference in MARL settings. 

\textbf{Causal discovery policy.} 
Our naive causal discovery approach exploits a random policy to build the causal graph. 
Obviously, that is inefficient as, generally speaking, there is no guarantee that the state-action-reward trajectories generated are informative enough with respect to the causal relationships to be learnt. 
A random policy also likely induce high variance, as demonstrated by the corresponding high variance of some of the results shown in Section~\ref{sec:eval}. 
A promising way forward is to use \emph{experimental design} principles to plan highly informative interventions (i.e.\ actions) in the environment yielding relevant causal information, as proposed by \cite{DBLP:conf/nips/KocaogluSB17}. 
Notice that our naive causal augmentation assumes that causal discovery precedes the RL stage, mostly for simplicity. 
Conceptually, though, nothing prevents an alternative view in which causal discovery and RL (there including the causal inference augmentation) proceed in parallel. 
In that case, deciding whether to use the same RL policy under training to guide causal discovery, or another arbitrary one, may be a though choice worth some formal and experimental analysis. 


\textbf{Causal model assessment.} 
The learnt causal model should be evaluated for correctness before being used for causal inference in the RL stage, as an incorrect model can worsen RL performance. 
However, validation of a learnt causal model is far from trivial, especially in MARL environments. 
First of all, the best way to assess the correctness of a causal model is to compare it to a \emph{ground truth} model. 
Though, in MARL, having access to the ground truth would translate to having access to a model of the agent-environment dynamics---making RL superfluous in the first place. 
%
Assessing a causal model without access to a ground truth, instead, is still an open issue---see \cite{DBLP:conf/icml/Karimi-Mamaghan24}. 
In our experiments, we simply blindly ``trust'' the learnt model, hence an incorrect one may explain the bad performance in some tasks---e.g.\ Give-Way. 

\textbf{Intervention on continuous variables.} 
The notion of \emph{intervention} is a cornerstone of the definition of causality given by \cite{DBLP:journals/ijon/Shanmugam01}. 
The core idea is to set the value of a variable to a specific one amongst the admissible domain. 
However, this conceptually simple mechanism may quickly become computationally intractable, especially when performing causal discovery. 
There, one may want to try \emph{every possible combination} of variable values to uncover potential causal effects on other variables. 
In the case of \emph{continuous variables} this brute-force mechanism is unfeasible. 
Discretisation offers a practical workaround to this issue, but it is not a principled solution, and may cause loss of relevant information. 
However, one promising research direction is to resort to \emph{soft interventions} (see \cite{DBLP:conf/nips/KocaogluJSB19}) in the form of distributional shifts, or ``tilts'' as defined in \cite{schindl2024incremental}, to probability density functions representing the domain of continuous variables. 

\textbf{Beyond the action filter.} 
In this paper, we used causal reasoning to learn a causal model to be used as an \emph{action space filter}. 
However, this is not the only way in which causal reasoning and (MA)RL can be integrated. 
Using counterfactuals for credit assignment, for instance, is another possible integration, as reported in Section~\ref{sec:back}. 
Furthermore, it is not straightforward to apply such filter in fully differentiable deep RL architectures, as the network is a black-box and the action pool cannot be easily filtered before action selection. 
Another challenge is balancing the RL algorithm's exploration-exploitation trade-off, which is designed to prevent overly greedy behaviour, with the ``guidance'' provided by the causal filter. 
On the one hand, one may want to rely as much as possible on the causal model to guide the agent in choosing actions ``rationally''; on the other hand, one must not overlap with the tasks for which RL is good---optimising the action policy. 

\textbf{Agents influences.} 
The naive causal augmentation approach that we put to test in this work is akin to an \emph{independent learning} setting in MARL: each agent attempts to learn a causal model of the MARL system in isolation, independently of other agents. 
As for MARL, this approach suffers from the problem of non-stationarity, as the dynamics to learn are influenced by learning agents in turn. 
We speculate that most of the failures in performance shown in Section~\ref{sec:eval} can be attributed to this limitation. 
Finding ways to foster collaboration amongst agents not only in the MARL stage, but during causal discovery, too, is likely to be key for learning correct and complete causal models. 
A promising research direction here started to appear in the pervasive systems and MAS domains, for instance by \cite{DBLP:conf/percom/MarianiZ24} and \cite{meganck2005distributed}. 
The former assume that computational nodes in a distributed environment cooperate to learn so-called ``minimal causal networks'' by coordinating interventions and exchanging data about variables' values probability distributions. 
The latter aims at letting agents in a MAS collaborate in building a global causal model by linking individual models via ``overlap variables''. 
%

\textbf{Formal account.} Besides experimental results, 
having formal proofs of convergence would be a cornerstone achievement. 
For causal MARL, and in particular for the kind of causal augmentation sketched in this paper, a first step would be to formalise the notion of what we called a ``minimal causal model'', or, perhaps equivalently, a causal model of the ``core environment dynamics''. 
Having a rigorous definition of what are the causal dynamics ``relevant'' for any given scenario-task pair would be a first step into a proper mathematical characterisation of causal MARL. 
It is worth mentioning two specific research works that, in our opinion, are going in this direction. 
\cite{DBLP:conf/aaai/MuttiSRCBR23} prove that, given assumptions on the ``common dynamics'' underlying a family of environments, perfect generalisation up to any desired error bound can be achieved via causal model-based RL. 
Such common dynamics may exactly be the one captured by a minimal causal model. 
\cite{liesen2024discovering} formulates a definition of a ``minimal RL environment'', linked to the improvement of learnt policy performance. Albeit not causal, this work can help shed light about how ``core'' environment dynamics may be defined. 

\section{Conclusions}
\label{sec:conc}

In this paper, we formulated a research roadmap shedding light on the open issues to deal with and promising directions to follow to successfully transfer the recent successful efforts in causal RL from a single-agent setting to MARL. 
To better ground our speculations, also informed by state-of-the-art literature, we performed experiments by augmenting popular MARL algorithms with causal discovery and inference, and testing the resulting causal versions on challenging MARL scenarios. 
The mixed results obtained, on the one hand show improvements even with the simple independent causal discovery approach adopted, on the other hand also exhibit limitations, especially in highly cooperative tasks. 
Based on this quantitative analysis, our roadmap details a few crucial steps that have to be made in the short term to advance the state of the art in causal MARL. 
In this context, our proposed causal augmentation is a first step on top of which the outlined roadmap can be developed.


\newpage
\appendix
\section{Codebase}

The code and execution instructions are provided in the submitted supplementary material. 
Documentation has been written for all project files to ease the reproduction process.

\subsection{Hyperparameters}
The experiment configurations follow the structure provided by the BenchMARL framework by \cite{DBLP:journals/jmlr/BettiniPM24}, utilizing Hydra (\cite{Yadan2019Hydra}) to separate YAML configuration files from the Python codebase. 


\section{Computational Resources Used}
This work required extensive computational resources. Specifically, for running the final experiments across multiple seeds (10, as recommended by \cite{DBLP:conf/nips/GorsaneMKDSP22}), we estimate approximately 400 hours of high-performance computing (HPC) time. The experiments were conducted using a server with an Intel(R) Xeon(R) CPU E5-2640 v4 @ 2.40GHz (20 cores) and 126GB of RAM, and an NVIDIA Titan X (Pascal) GPU with 12GB of RAM. 

\section{Computational Complexity}
In this section, we focus on analyzing the computational complexity of causal discovery and inference, specifically of building the Causal Bayesian Network (CBN) used to incorporate causal knowledge. 


\subsection{Causal Discovery}
Causal discovery involves learning the graph structure, and we use the PC algorithm by \cite{DBLP:books/daglib/0023012}, a constraint-based method with complexity based on two main phases:
\begin{enumerate}
    \item \textit{Skeleton Construction.} The algorithm performs conditional independence tests to decide if an edge exists between pairs of variables. In the worst case, the complexity is $\mathcal{O}(p^d 2^d)$, where \( p \) is the number of variables (nodes) and \( d \) is the maximum degree (edges per node).

   \item \textit{Edge Orientation.} Once the skeleton (undirected graph) is built, edges are oriented using v-structures, with a complexity of \( \mathcal{O}(p^2) \).
\end{enumerate}

\subsection{Causal Inference}
Once the graph is constructed, interventions are computed by manipulating the joint probability distribution over the nodes in the Markov blanket of the reward. 
The complexity depends on the number of variables and their interdependencies within the blanket. 
Generally, exact inference in a CBN can be $exponential$ in the network size, especially for more complex structures. 


\newpage
\section{Architecture of Causal MARL Algorithms}
\label{app:architecture_causal_benchmarl}

Figure~\ref{alg:causal_forward_pass} shows the actual architecture of the causal augmentation of the vanilla algorithms chosen. 

\begin{figure}[!ht]
    \centering
    {%
        \includegraphics[width=0.7\columnwidth]{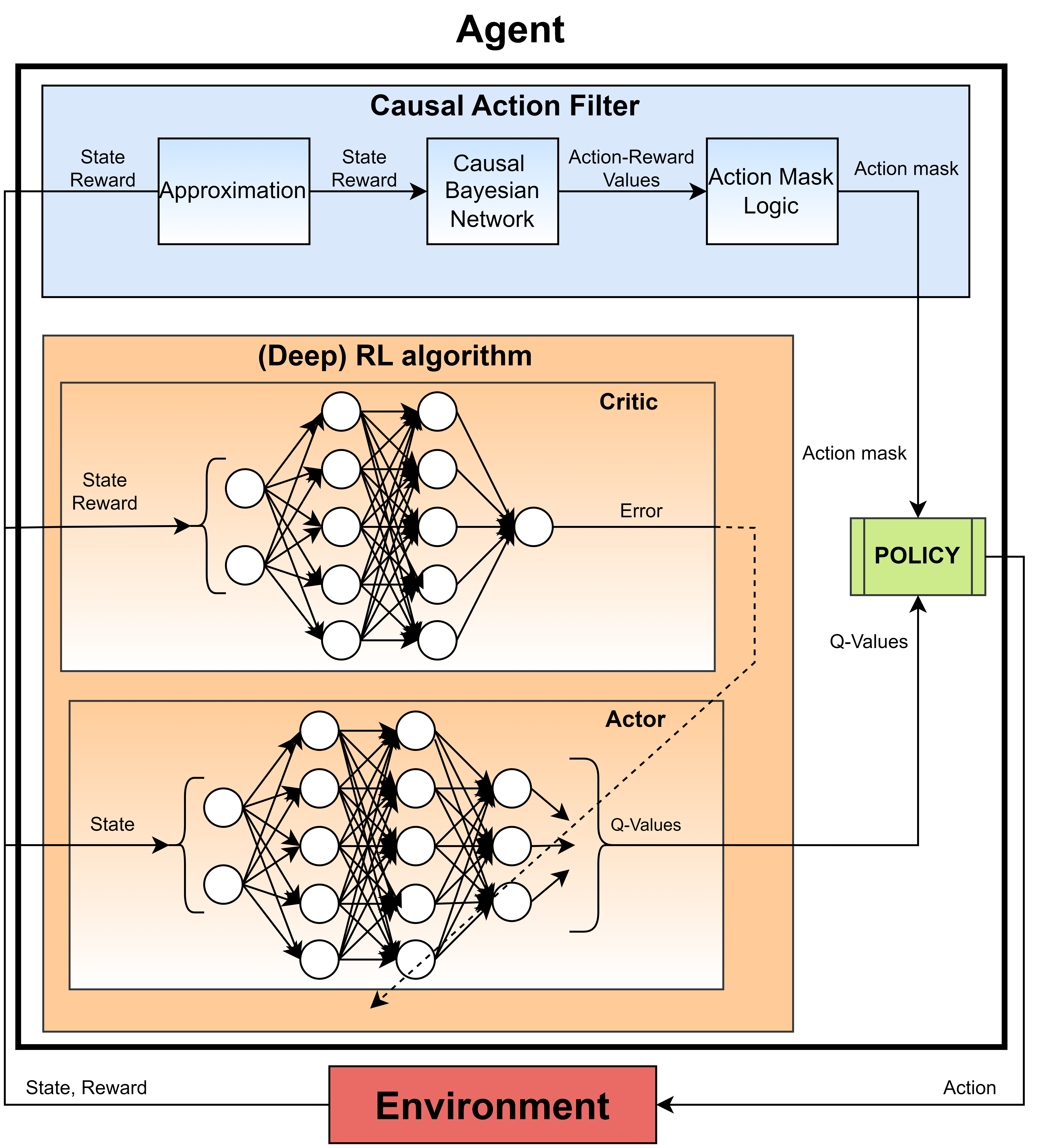}
    }
    {\caption{Specific CDRL architecture employed for BenchMARL algorithms: the policy considers both the actor-critic output, similar to the baselines, as well as the actions that are selectable and those that are not, based on the causal action mask. The key portion of this process, the forward pass, is detailed in Algorithm \ref{alg:causal_forward_pass}.}}
    \label{fig:causal_benchmarl}
\end{figure}

\newpage
\section{Sensitive Analysis}
\label{app:sensitive}

In this section, we outline the sensitivity analysis carried out to identify the optimal set of approximation parameters (e.g.\ discretisation bins and method). 

\emph{VMAS} by \cite{DBLP:conf/dars/BettiniKBP22} is a vectorized 2D physics engine in \texttt{PyTorch} with challenging multi-robot scenarios, where: the observation space is continuous, with different size based on the task under consideration, and the action space can be discrete or continuous, based on the user choice. 

In this work, we evaluate our Causality-Driven RL framework across the navigation, flocking, and give-way scenarios, as shown in Figure \ref{fig:VMAS_envs}. 
We rely on discrete actions (provided by VMAS), discrete variables, and a limited observation space. 
For tasks involving Lidar sensor (navigation and flocking), we simplify the information to reduce the number of features. Specifically, for a given $N$ number of Lidar sensors, we use the sensor with the maximum value and its corresponding value as features, and so on, resulting in two features per sensor. 
Once the observation space is reduced, we discretize all features into $M$ bins.


To summarize, the parameters we are investigating through this sensitivity analysis are:
\begin{itemize}
    \item $L$: the number of samples required to acquire accurate causal knowledge
    \item $M$: the number of bins needed to discretize the observation space without losing significant information
    \item $N$: the number of sensors to consider that best capture the collision dynamics
\end{itemize}

The policy for acquiring samples during this process is entirely random. To reduce exploration inefficiencies, we constrained the environment's size in scenarios where it wasn't already limited (e.g., the navigation scenario). 

\begin{figure}[!ht]
    \centering
    {%
        \includegraphics[width=\textwidth]{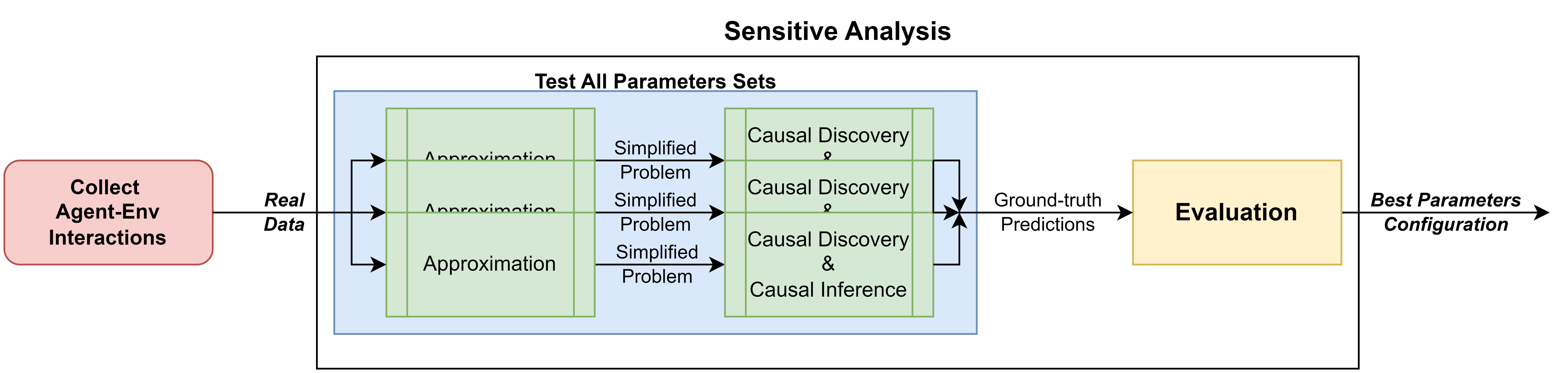}
    }
    \caption{Sensitive analysis architecture summary.}
    \label{fig:sensitive_analysis} 
\end{figure}

The sensitivity analysis we developed is structured as shown in Figure \ref{fig:sensitive_analysis}. Specifically, two key tasks are considered:

\begin{enumerate}
    \item \textbf{Causal Discovery}: This task involves learning the causal structure, represented as a Directed Acyclic Graph (DAG), from data. To evaluate the quality of the learned DAG 
    we use several graph-based metrics, which are grouped into two categories: 
    \begin{enumerate} 
        \item \textit{Graph-distance metrics}: Structural Hamming Distance, Structural Intervention Distance, Frobenius Norm. 
        \item \textit{Graph-similarity metrics}: Jaccard Similarity, Degree Distribution Similarity and Clustering Coefficient Similarity.
    \end{enumerate}

    Given the large observation space and continuous variables within, applying the PC algorithm \cite{DBLP:books/daglib/0023012} directly would be computationally expensive and challenging. 
    Therefore, we approximate the data using parameters $L=10^6$, $M=100$, and $N=4$, perform causal discovery on this approximated dataset, and treat the resulting DAG as the ground truth.

    \item \textbf{Causal Inference}: This task focuses on analyzing the response of an effect variable when its cause is modified. Specifically, we aim to predict the value of the ``reward'' variable based on the observation and the action taken. For the ground truth, we use the original continuous data collected from VMAS, while the ``predicted values'' are generated from a Causal Bayesian Network trained on the approximated data and the learned DAG from that data. 
    To assess the quality of our predictions, we apply several value-distance metrics, grouped into two categories:
    \begin{itemize}
    \item \textit{Binary-distance metrics}: Accuracy, Precision, F1-score, Recall.
    \item \textit{Value-distance metrics}: Mean Absolute Error, Mean Squared Error, Root Mean Square Error, Median Absolute Error.
    \end{itemize}
\end{enumerate}

We rescale the results of each metric to a range between 0 and 1 (in the causal-distance metrics category, we use a fully connected graph as an edge case) and adjust the values so that the maximum value is set to 1. 
Then, by averaging the metrics within each category, we obtain four values -- one for each category -- for every combination of sensitivity analysis parameters. 
To select the optimal configuration, we multiply the averaged values from each metric category. 
The best configuration is the one with the highest final value.

Here, we present the set of sensitivity parameters tested for each task and the best parameters configurations:
\begin{itemize}
    \item \textbf{Navigation}:
    \begin{itemize}
        \item $L$: [500000, 1000000];
        \item $M$: [10, 15, 20, 25];
        \item $N$: [1, 2];
        \item $Best$: $L$=1000000 and $M$=10 and $N$=2. Score: 0.146.
    \end{itemize}

    \item \textbf{Flocking}:
    \begin{itemize}
        \item $L$: [500000, 1000000];
        \item $M$: [10, 15, 20, 25];
        \item $N$: [1, 2];
        \item $Best$: $L$=500000 and $M$=10 and $N$=1. Score 0.444.
    \end{itemize}

    \item \textbf{Give-way}:
    \begin{itemize}
        \item $L$: [500000, 1000000];
        \item $M$: [10, 15, 20, 25];
        \item $Best$: $L$=500000 and $M$=10. Score: 0.0 (selected due to reduced computational expense).
    \end{itemize}
\end{itemize}

\begin{figure}[!b]
    \centering
    {%
        \begin{subfigure}[]{.2\columnwidth}
            \includegraphics[width=\columnwidth]{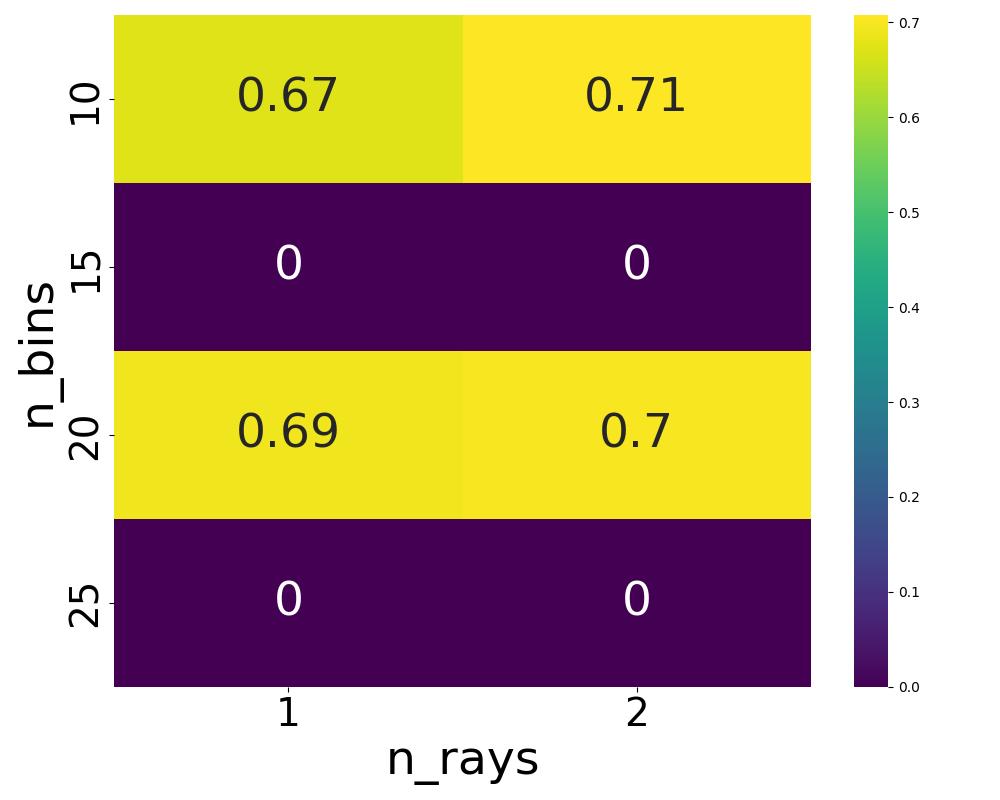}
            \caption{Binary-distance metrics}
            \label{fig:navigation_binary_metrics}
        \end{subfigure}
        \hspace{0.01\textwidth}
        \begin{subfigure}[]{.2\columnwidth}
            \includegraphics[width=\columnwidth]{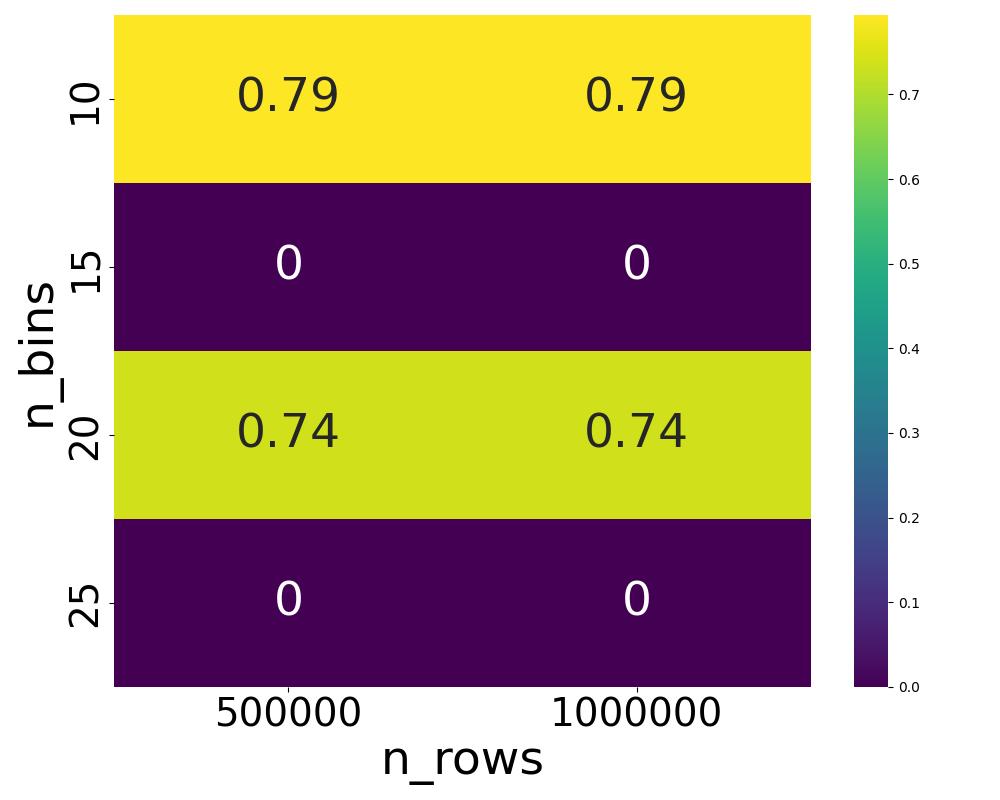}
            \caption{Value-distance metrics}
            \label{fig:navigation_distance_metrics}
        \end{subfigure}
        \hspace{0.01\textwidth}
        \begin{subfigure}[]{.2\columnwidth}
            \includegraphics[width=\columnwidth]{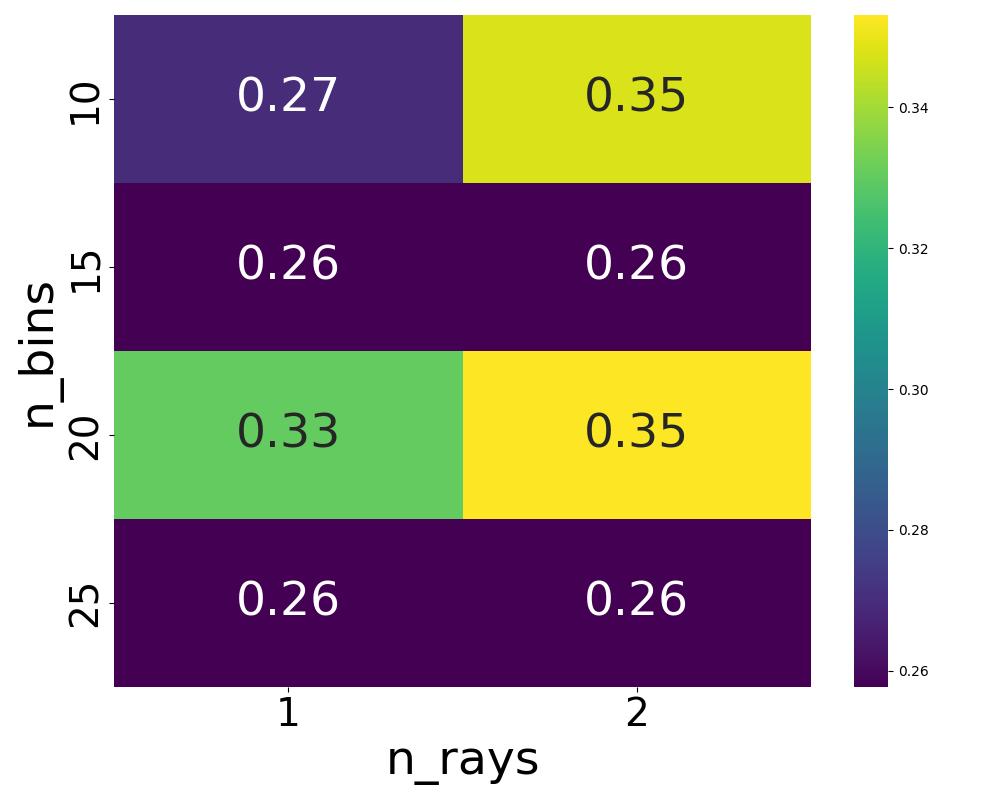}
            \caption{Graph similarity metrics}
            \label{fig:navigation_causal_graph_similarity_metrics}
        \end{subfigure}
        \hspace{0.01\textwidth}
        \begin{subfigure}[]{.2\columnwidth}
            \includegraphics[width=\columnwidth]{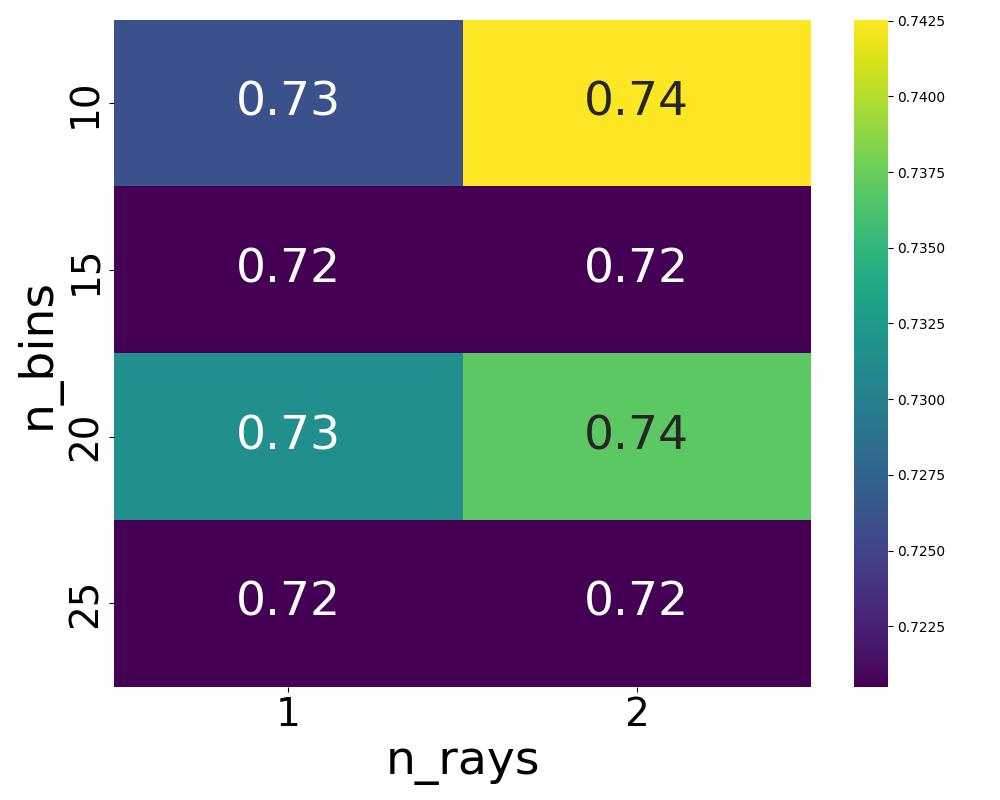}
            \caption{Graph distance metrics}
            \label{fig:navigation_causal_graph_distance_metrics}
        \end{subfigure}
    }
    \caption{Sensitive analysis for the Navigation task.}
    \label{fig:navigation_sensitive_analysis} 
\end{figure}

\begin{figure}[!b]
    \centering
    {%
        \begin{subfigure}[]{.2\columnwidth}
            \includegraphics[width=\columnwidth]{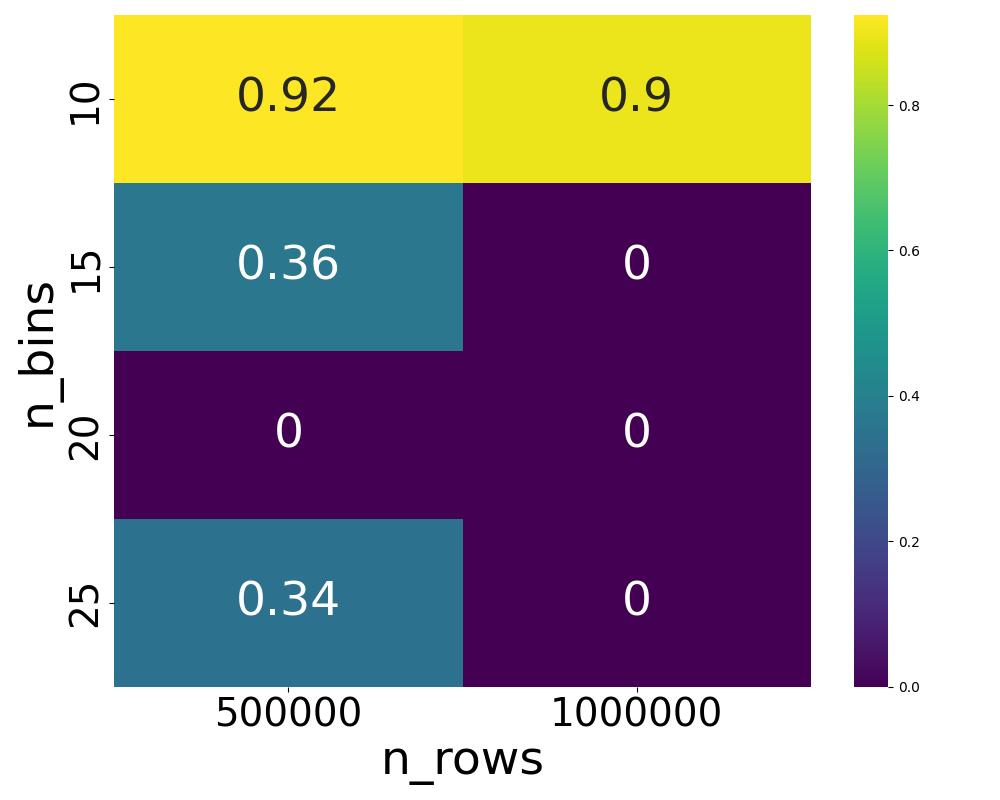}
            \caption{Binary-distance metrics}
            \label{fig:flocking_binary_metrics}
        \end{subfigure}
        \hspace{0.01\textwidth}
        \begin{subfigure}[]{.2\columnwidth}
            \includegraphics[width=\columnwidth]{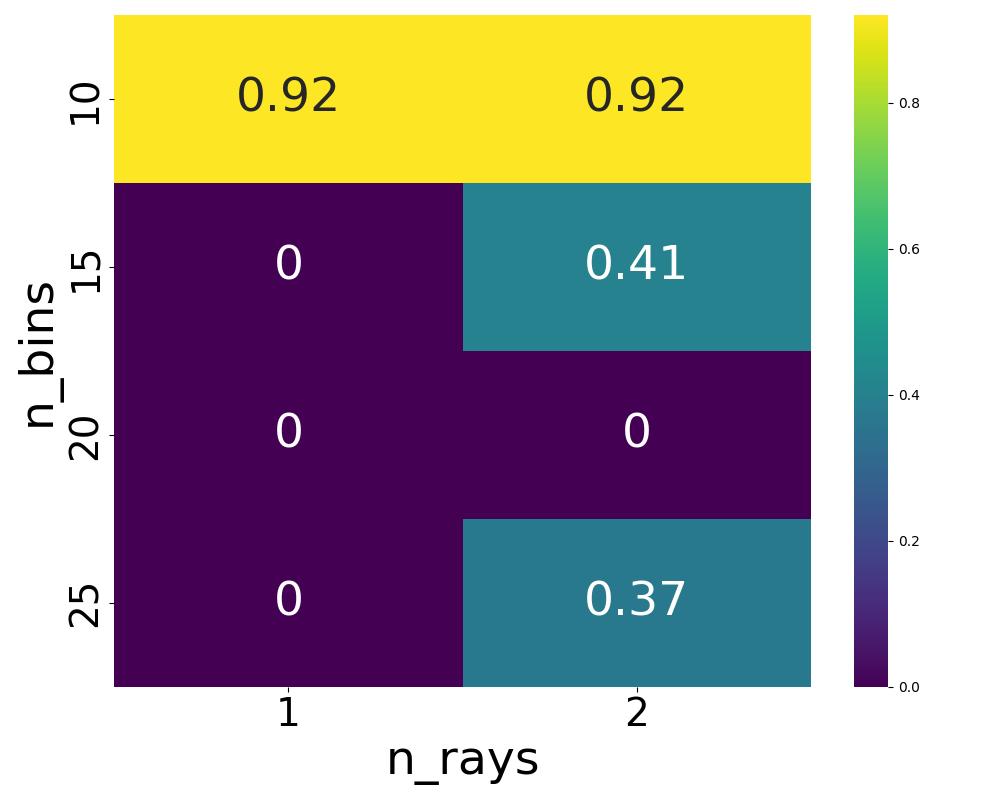}
            \caption{Value-distance metrics}
            \label{fig:flocking_distance_metrics}
        \end{subfigure}
        \hspace{0.01\textwidth}
        \begin{subfigure}[]{.2\columnwidth}
            \includegraphics[width=\columnwidth]{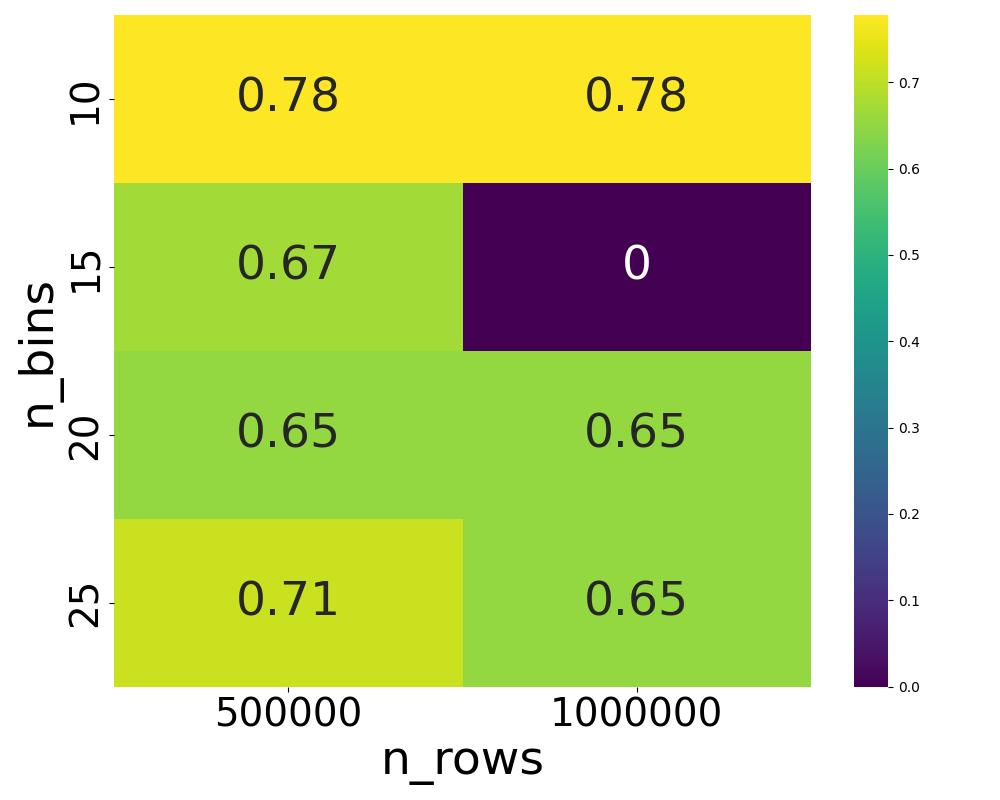}
            \caption{Graph similarity metrics}
            \label{fig:flocking_causal_graph_similarity_metrics}
        \end{subfigure}
        \hspace{0.01\textwidth}
        \begin{subfigure}[]{.2\columnwidth}
            \includegraphics[width=\columnwidth]{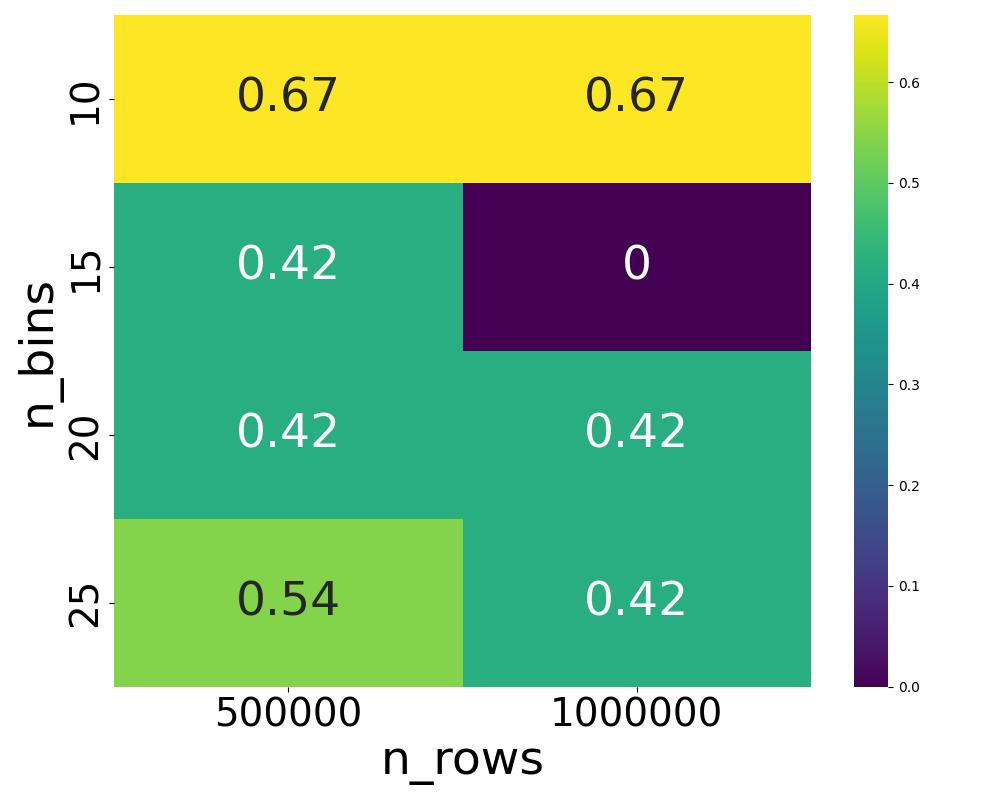}
            \caption{Graph distance metrics}
            \label{fig:flocking_causal_graph_distance_metrics}
        \end{subfigure}
    }
    \caption{Sensitive analysis for the Flocking task.}
    \label{fig:flocking_sensitive_analysis} 
\end{figure}

\begin{figure}[!t]
    \centering
    {%
        \begin{subfigure}[]{.2\columnwidth}
            \includegraphics[width=\columnwidth]{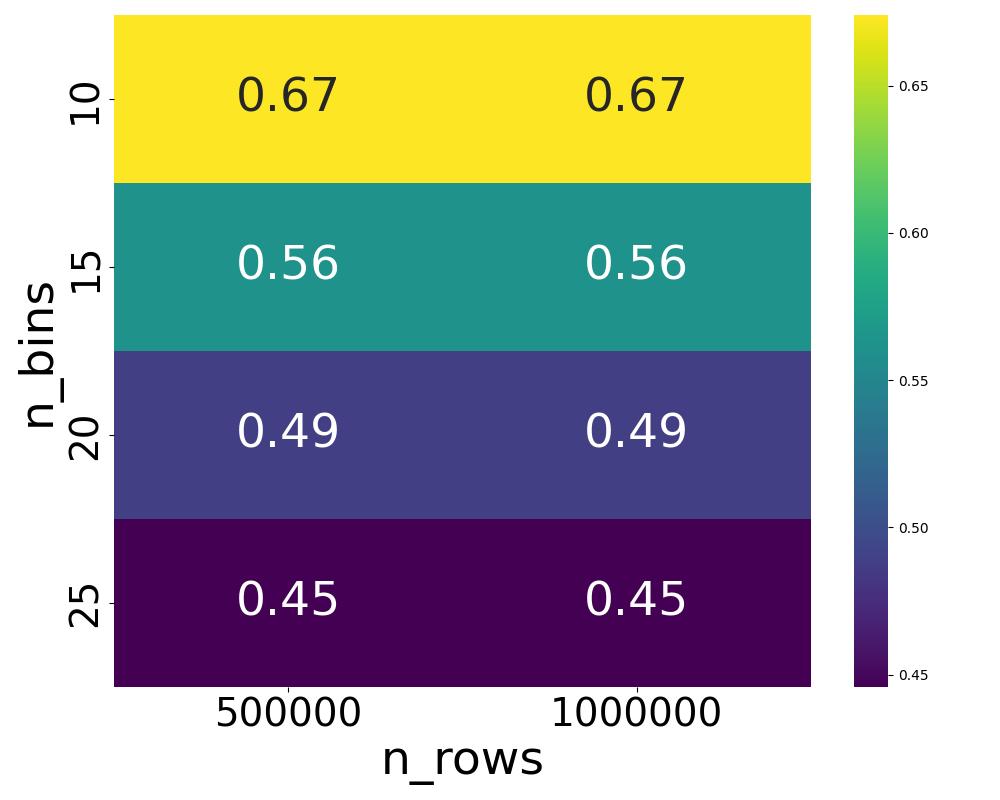}
            \caption{Binary-distance metrics}
            \label{fig:giveway_binary_metrics}
        \end{subfigure}
        \hspace{0.01\textwidth}
        \begin{subfigure}[]{.2\columnwidth}
            \includegraphics[width=\columnwidth]{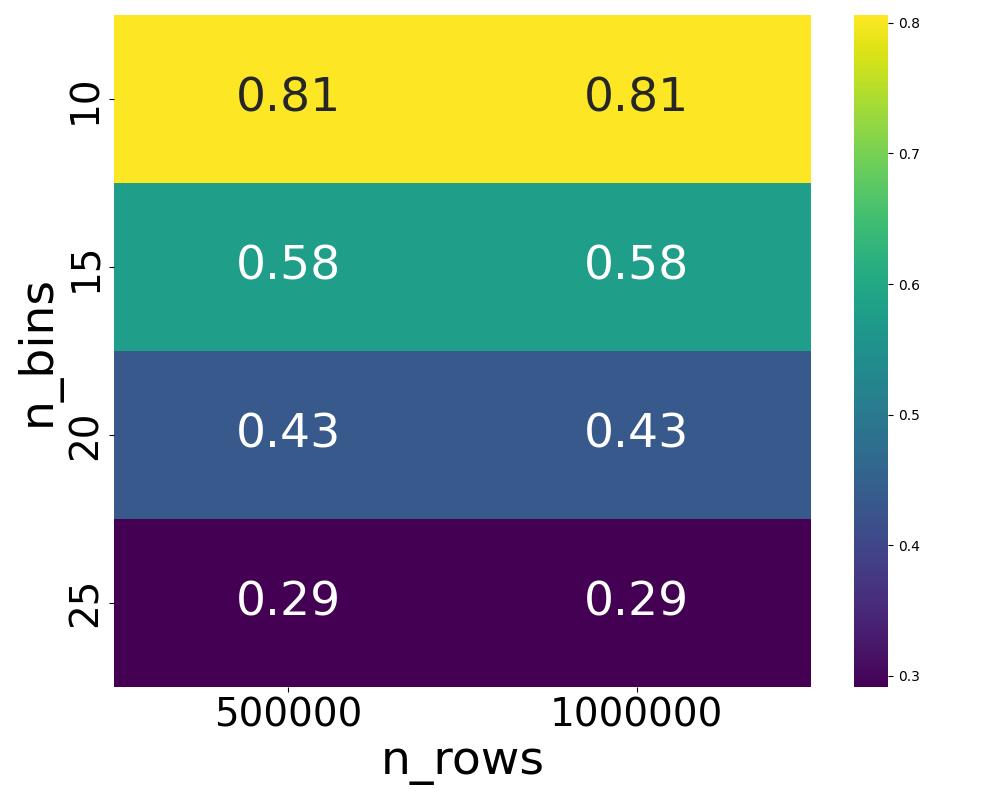}
            \caption{Value-distance metrics}
            \label{fig:giveway_distance_metrics}
        \end{subfigure}
        \hspace{0.01\textwidth}
        \begin{subfigure}[]{.2\columnwidth}
            \includegraphics[width=\columnwidth]{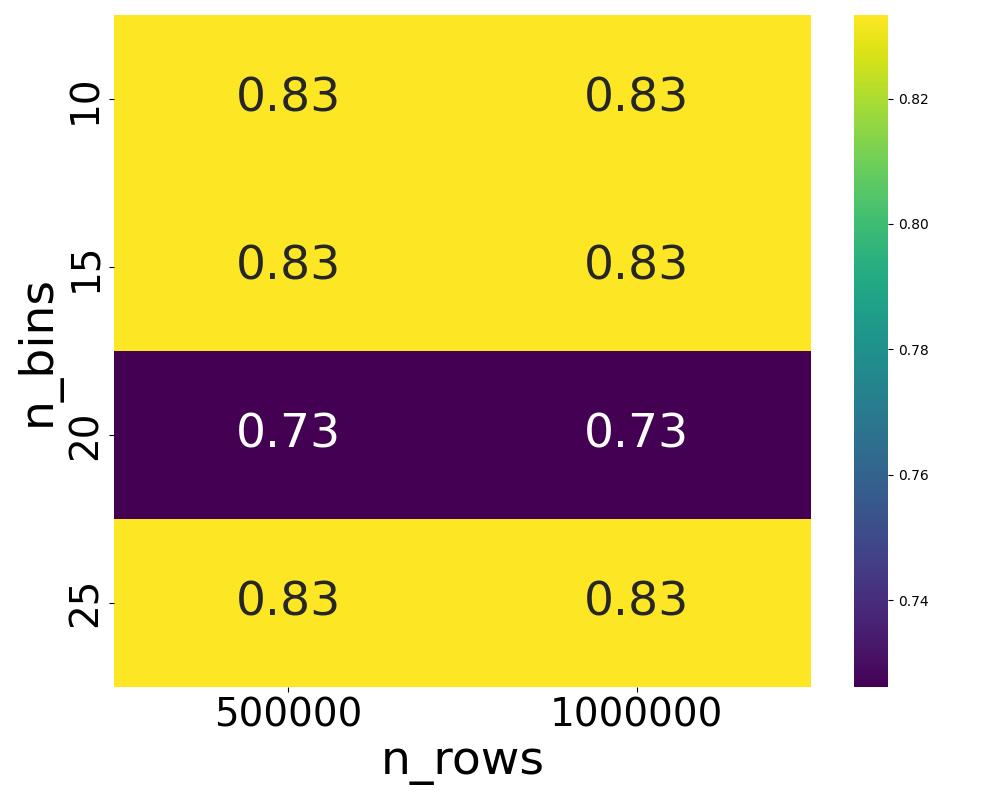}
            \caption{Graph similarity metrics}
            \label{fig:giveway_causal_graph_similarity_metrics}
        \end{subfigure}
        \hspace{0.01\textwidth}
        \begin{subfigure}[]{.2\columnwidth}
            \includegraphics[width=\columnwidth]{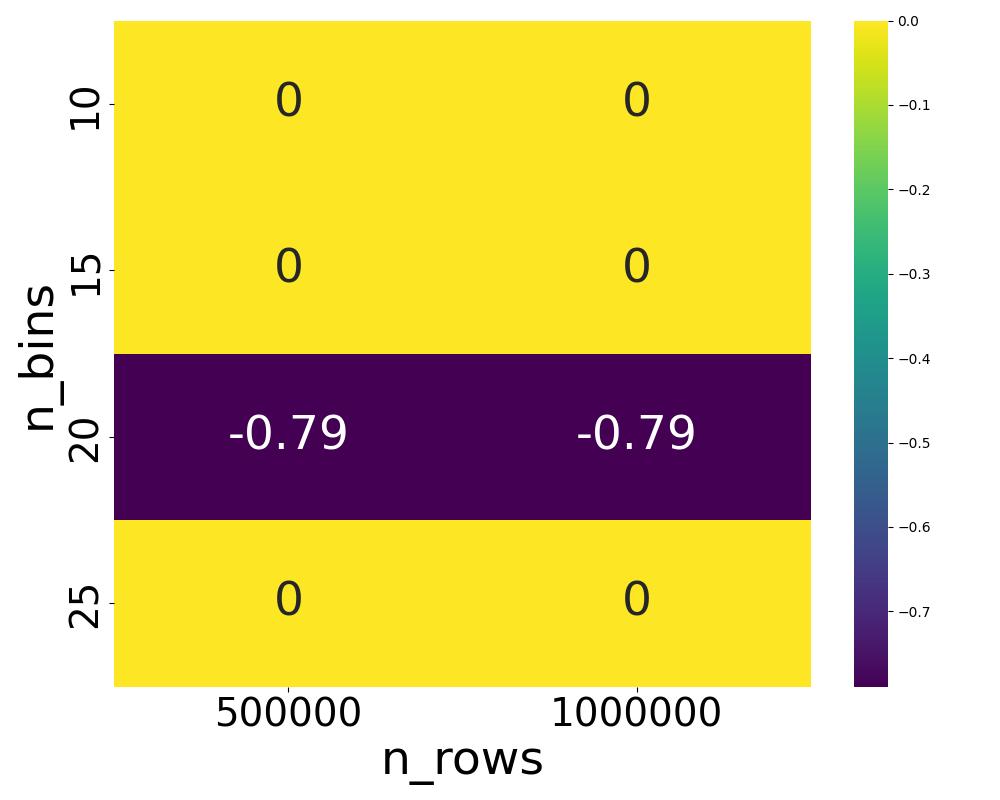}
            \caption{Graph distance metrics}
            \label{fig:giveway_causal_graph_distance_metrics}
        \end{subfigure}
    }
    \caption{Sensitive analysis for the Give-Way task.}
    \label{fig:giveway_sensitive_analysis} 
\end{figure}

Figures \ref{fig:navigation_sensitive_analysis}, \ref{fig:flocking_sensitive_analysis}, and \ref{fig:giveway_sensitive_analysis} present the confusion matrices for each configuration of sensitivity parameters, organized by task. In these plots, only two parameters are displayed along the axes. For tasks with three parameters (navigation and flocking), the two most significant parameters were selected for display based on an importance analysis performed using a \texttt{RandomForestRegressor}.

\section{Learnt Causal DAGs}
In this section, we present the extracted DAGs for each task, based on the best approximations identified through the sensitivity analysis outlined earlier. 

\begin{figure}[!ht]
    \centering
    {%
        \begin{subfigure}[]{.45\columnwidth}
            \includegraphics[width=\columnwidth]{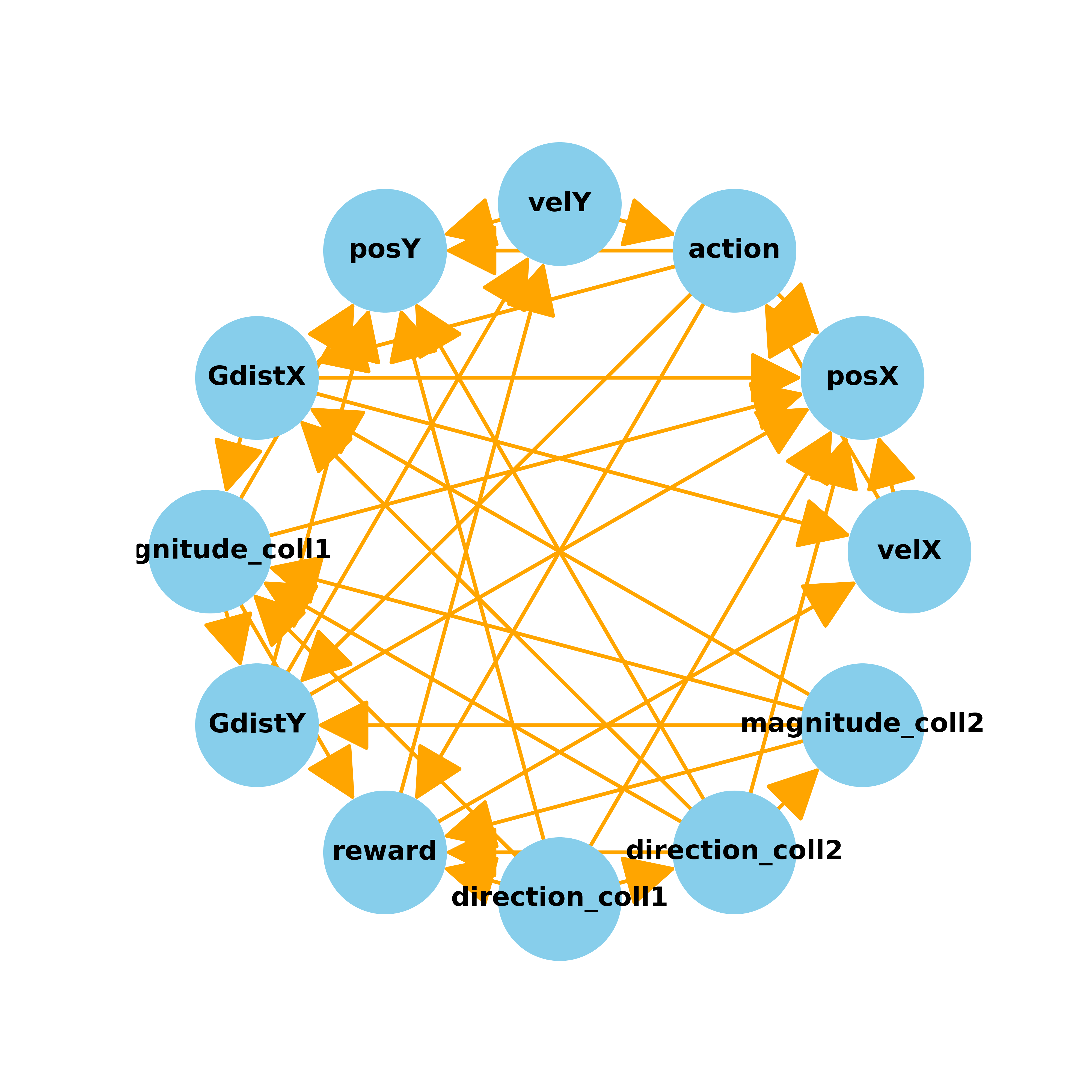}
            \caption{Complete DAG}
            \label{fig:navigation_complete_dag}
        \end{subfigure}
        \hspace{0.02\textwidth}
        \begin{subfigure}[]{.45\columnwidth}
            \includegraphics[width=\columnwidth]{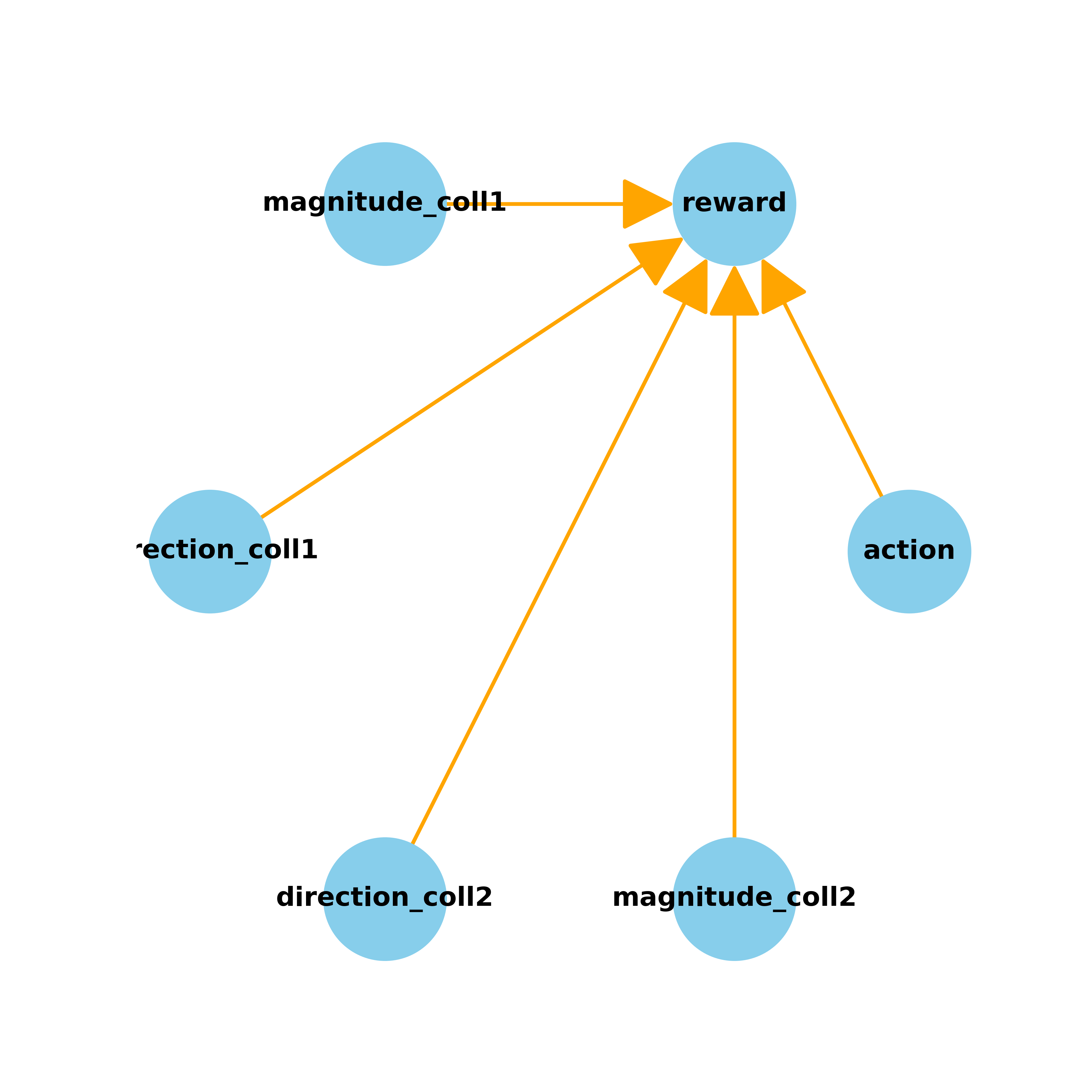}
            \caption{Minimal DAG}
            \label{fig:navigation_minimal_dag}
        \end{subfigure}
    }
    {\caption{Navigation scenario: \textit{Left}, the complete DAG outcomes from Causal Discovery. \textit{Right}, the minimal DAG considered after our assumptions.}} 
    \label{fig:navigation_dags} 
\end{figure}

\begin{figure}[!ht]
    \centering
    {%
        \begin{subfigure}[]{.45\columnwidth}
            \includegraphics[width=\columnwidth]{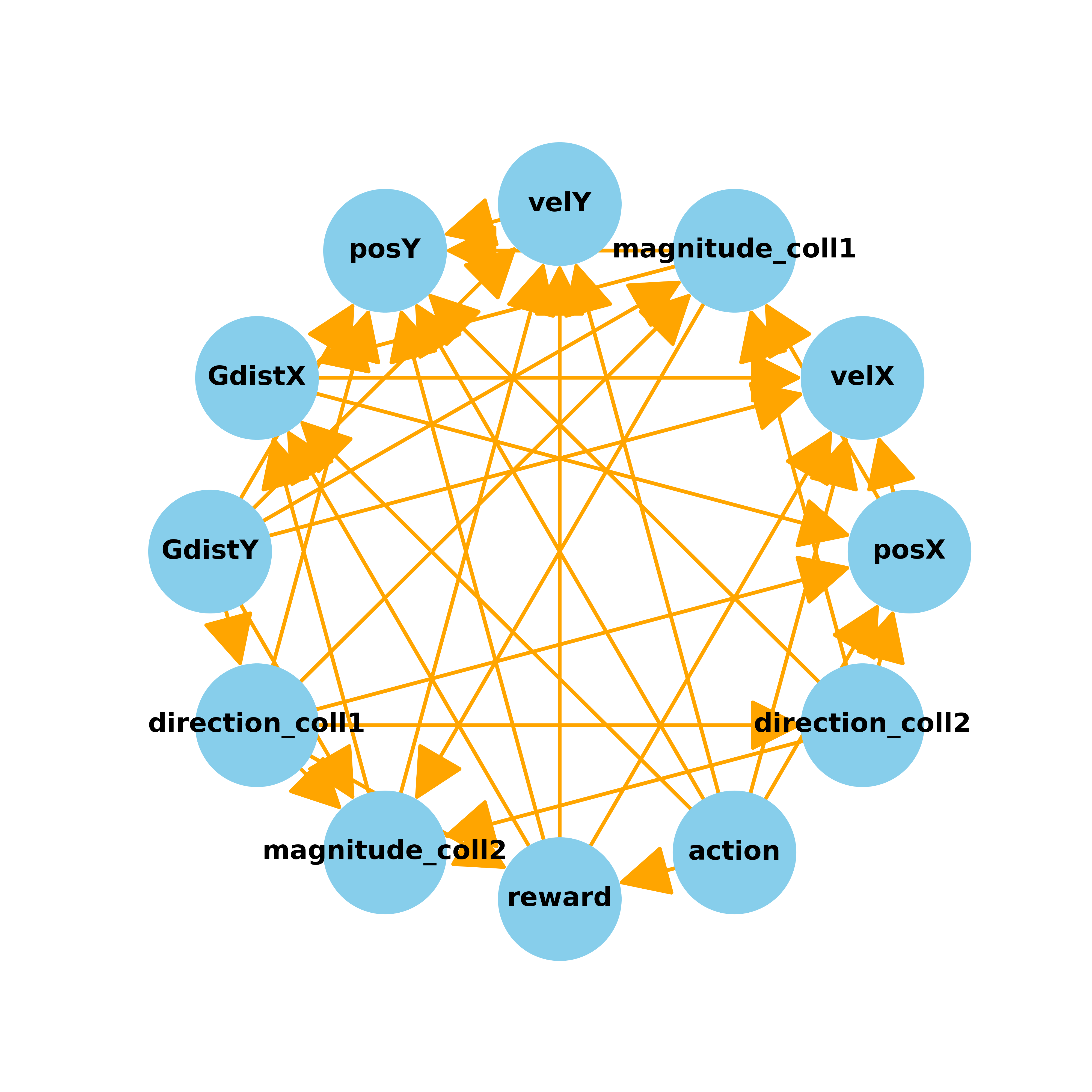}
            \caption{Complete DAG}
            \label{fig:flocking_complete_dag}
        \end{subfigure}
        \hspace{0.02\textwidth}
        \begin{subfigure}[]{.45\columnwidth}
            \includegraphics[width=\columnwidth]{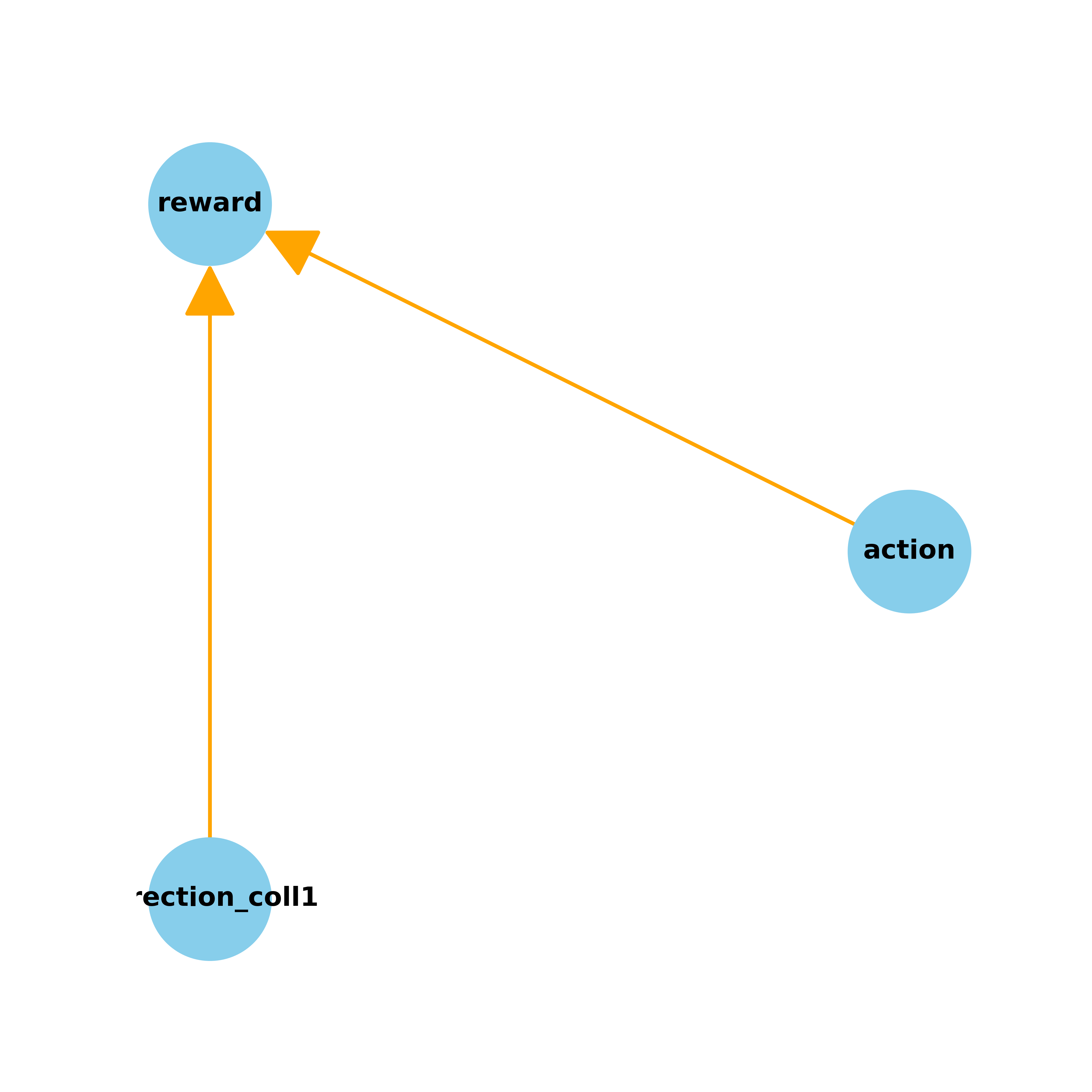}
            \caption{Minimal DAG}
            \label{fig:flocking_minimal_dag}
        \end{subfigure}
    }
    {\caption{Flocking scenario: \textit{Left} shows the complete DAG from Causal Discovery, while \textit{Right} displays the minimal DAG after applying our assumptions.}} 
    \label{fig:flocking_dags} 
\end{figure}

\begin{figure}[!ht]
    \centering
    {%
        \begin{subfigure}[]{.45\columnwidth}
            \includegraphics[width=\columnwidth]{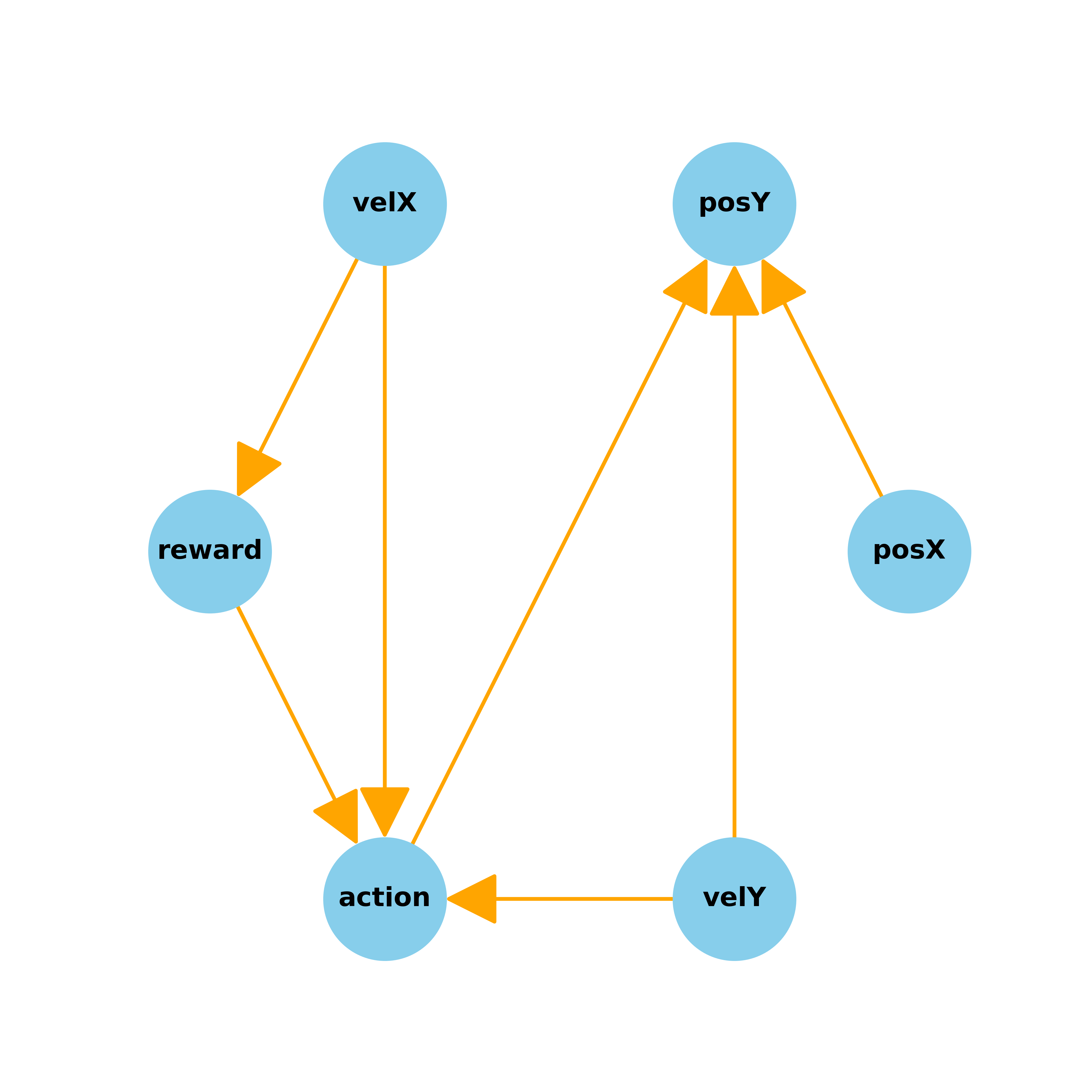}
            \caption{Complete DAG}
            \label{fig:give_way_complete_dag}
        \end{subfigure}
        \begin{subfigure}[]{.45\columnwidth}
            \includegraphics[width=\columnwidth]{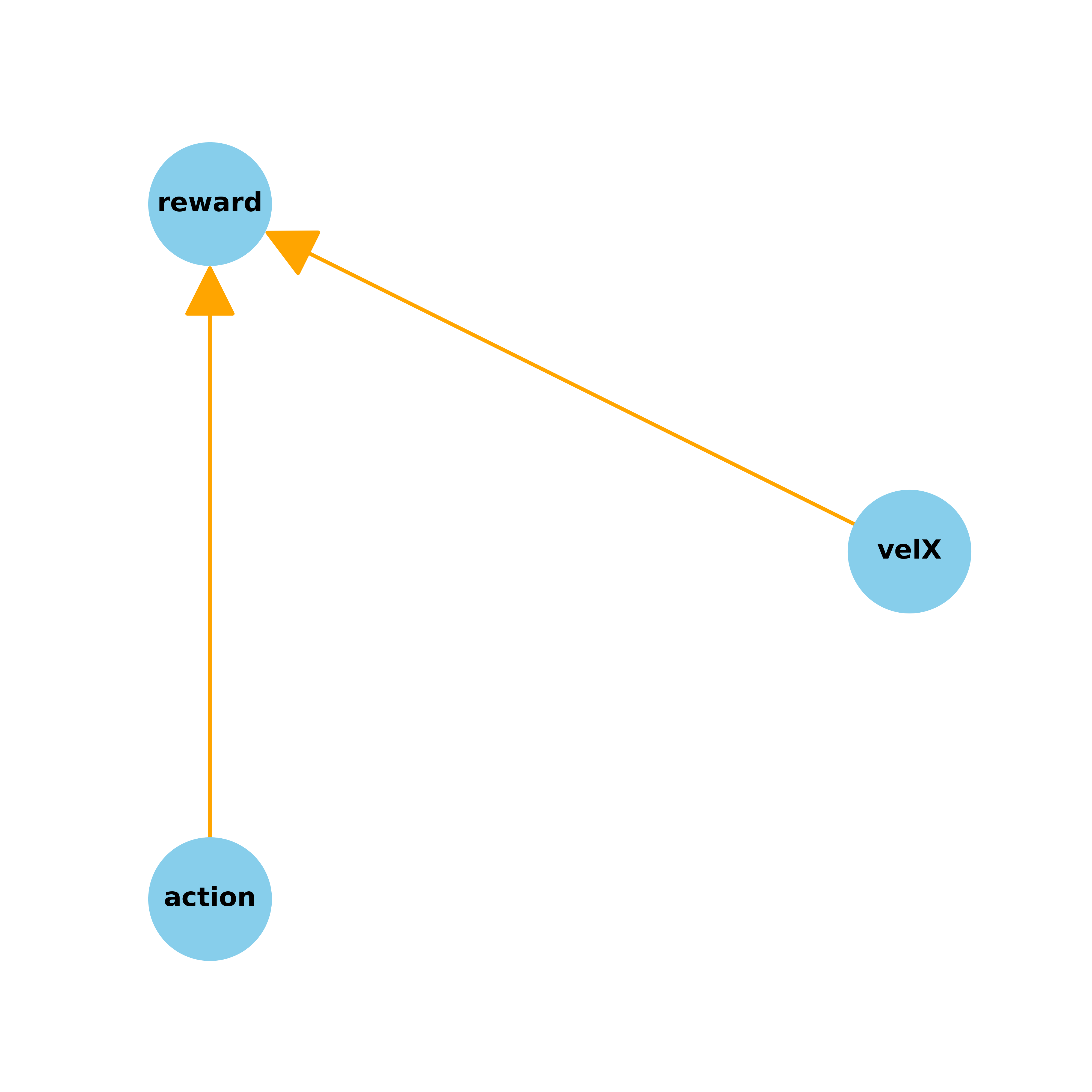}
            \caption{Minimal DAG}
            \label{fig:give_way_minimal_dag}
        \end{subfigure}
    }
    \label{fig:give_way_dags} 
    {\caption{Give-Way scenario: \textit{Left} shows the complete DAG from Causal Discovery, while \textit{Right} displays the minimal DAG after applying our assumptions.}} 
\end{figure}

\end{document}